\documentclass{article}

\usepackage{microtype}
\usepackage{graphicx}
\usepackage{subcaption}
\usepackage{booktabs} 

\usepackage{hyperref}

\usepackage[accepted]{icml2026}

\usepackage{amsmath}
\usepackage{amssymb}
\usepackage{mathtools}
\usepackage{amsthm}

\usepackage{enumitem}
\usepackage{multirow}
\usepackage{tcolorbox}
\usepackage{adjustbox}
\usepackage{pifont}

\usepackage{xcolor}


\newcommand\model[1]{{\footnotesize \texttt{#1}}}

\newcommand\deepseekLlama{\model{DeepSeek-R1-Distill-Llama-8B}}
\newcommand\deepseekQwen{\model{DeepSeek-R1-Distill-Qwen-14B}}
\newcommand\QwQLarge{\model{QwQ-32B}}

\newcommand{\cmark}{\ding{51}} 

\usepackage[capitalize,noabbrev]{cleveref}

\theoremstyle{plain}

\theoremstyle{definition}

\theoremstyle{remark}

\usepackage[textsize=tiny]{todonotes}

\icmltitlerunning{Step-Tagging: Toward controlling the generation of
Language Reasoning Models through black-box step monitoring}

\begin{document}

\twocolumn[
  \icmltitle{Step-Tagging Early-Stopping: Toward controlling the generation of \\ Language Reasoning Models through black-box step monitoring}



  \icmlsetsymbol{equal}{*}

  \begin{icmlauthorlist}
    \icmlauthor{Yannis Belkhiter}{comp,sch}
    \icmlauthor{Seshu Tirupathi}{comp}
    \icmlauthor{Giulio Zizzo}{comp}
    \icmlauthor{John D. Kelleher}{yyy,sch}
  \end{icmlauthorlist}

  \icmlaffiliation{sch}{Trinity College Dublin, Ireland}
  \icmlaffiliation{comp}{IBM Research Europe, Dublin, Ireland}
  \icmlaffiliation{yyy}{ADAPT Research Centre, Dublin, Ireland}

  \icmlcorrespondingauthor{Yannis Belkhiter}{yannis.belkhiter@ibm.com}

  \icmlkeywords{Machine Learning, ICML}

  \vskip 0.3in
]



\printAffiliationsAndNotice{}  

\begin{abstract}
  Language Reasoning Models (LRMs) have shown impressive performance on solving complex problems requiring multi-steps. However, a growing body of studies show that LRMs are still inefficient, over-generating verification and self-reflection steps. To address this challenge, we introduce the Step-Tagging Early-Stopping (ST-ES) framework, a lightweight sentence-classifier enabling real-time annotation of the type of reasoning steps that an LRM is generating. We show that limiting the count of specific step-type - especially verification and self-reflection steps - yields a more accurate and token-efficient early-stopping criterion than token-count baseline, and that each step-types yield to a different efficiency trade-off. Unlike prior dynamic early-stopping methods, ST-ES operates in a full black-box setting, and offers interpretable early-stopping criteria. We evaluate ST-ES on three mathematical reasoning benchmarks, namely, MATH500, GSM8K, AIME and two knowledge and reasoning benchmarks, MMLU and GPQA respectively. We achieve 20 to 50\% token reduction while maintaining comparable accuracy to standard generation.

\end{abstract}

\section{Introduction}

For the past few years, the field of Language Reasoning Models (LRMs) has experienced significant growth in terms of capabilities. 
Inference Time Scaling, with work on model prompting has emerged as a popular field with the goal of making models more accurate at reasoning \citep{wei2023chainofthoughtpromptingelicitsreasoning, wang2023selfconsistencyimproveschainthought}. At the same time, fundamental work on Training Time Scaling has led to the release of strong native reasoning models \cite{xu2025largereasoningmodelssurvey}. \\
\phantom{m} 
However, recent surveys have shown that LRMs need to generate a very large number of tokens---several thousands---in order to generate an accurate answer on challenging questions \citep{qu2025surveyefficientreasoninglarge, chen2025think23overthinkingo1like, sui2025stopoverthinkingsurveyefficient}. This behavior makes reasoning models extremely inefficient - scaling in both compute resources and inference time. Although recent works have suggested solutions to this problem, most approaches overlook the possibility of monitoring the reasoning as it is being generated, and using this information to improve the inference such as its efficiency. Indeed, efficient inference-time scaling methods are either static - fixed budgets/black-box - or dynamic, but they requires access the model's internals and do not exploit the text generated by LRMs to guide early stopping. 
To address this challenge, this paper aims to offer a new perspective on the efficiency of LRMs by focusing on dynamic monitoring of model's outputs. Our contributions are as follows:

\begin{figure}[t]
    \centering
    \includegraphics[width=1.0\linewidth]{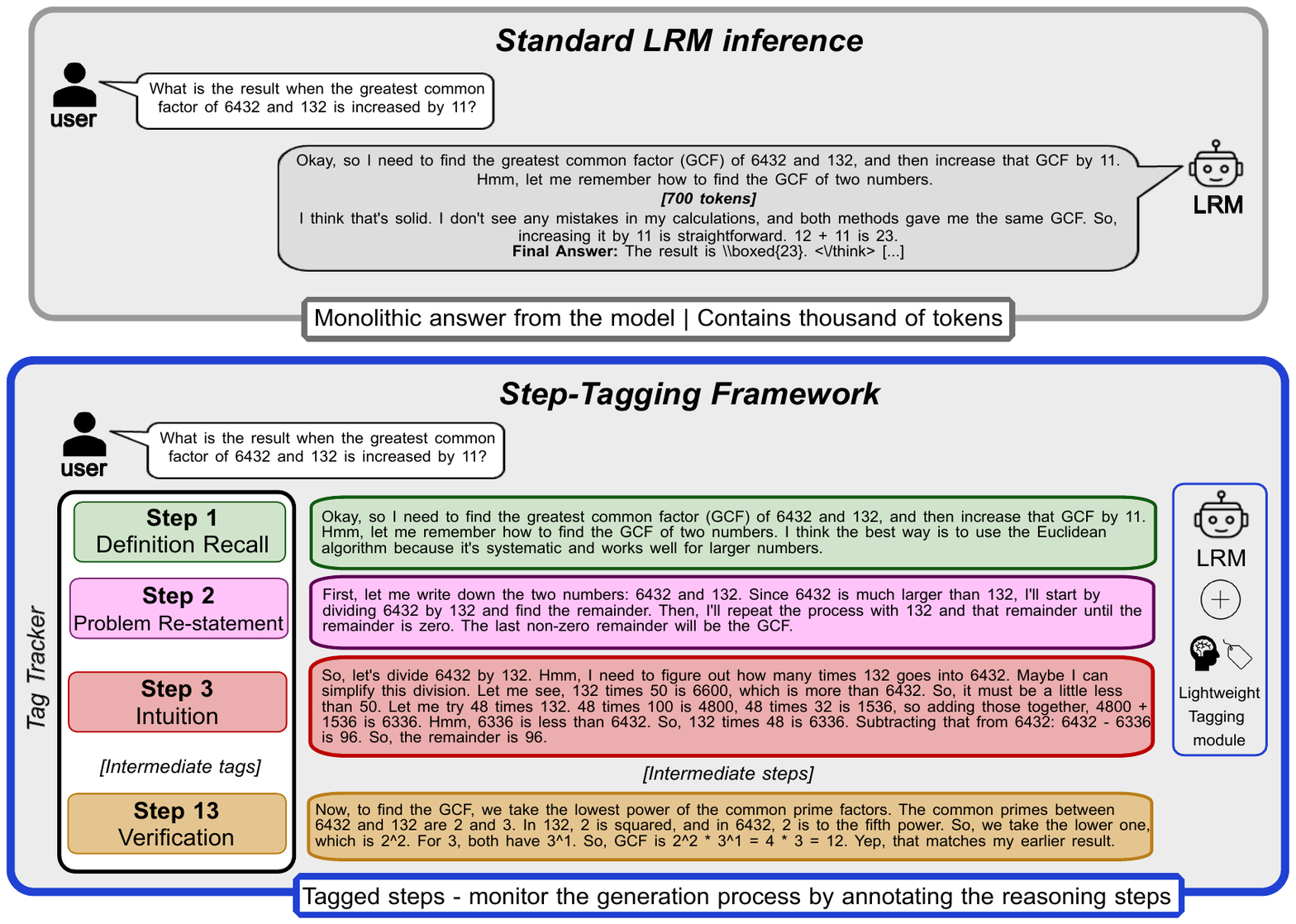}
    \vspace{-0.35cm}
    \caption{\small Step-Tagging: a framework for monitoring the generation of LRMs - example on sample 39 from MATH500 test with DS-Qwen14B, using the ReasonType taxonomy - seed $42$}
    \label{fig:step-tagging}
    \vspace{-0.45cm}
\end{figure}

\begin{itemize}[leftmargin=*, itemsep=0pt, topsep=0pt]

    \item \textbf{Step-Tagging module:} We introduce the \emph{Step-Tagging} module (see Figure \ref{fig:step-tagging}), an online lightweight sentence classifier capable of identifying the nature of each step generated by LRMs. This novel framework offers a tool to systematically monitor the generation of LRMs. 

    \item \textbf{Step-Tagging Early-Stopping (ST-ES):} Leveraging the \emph{Step-Tagging} module, we found that the type of reasoning steps plays a role in determining the early-stopping condition. Based on these observations, we built an interpretable early-stopping framework that dynamically stops token generation based on reasoning steps types and counts, calibrated on both models and problem complexity. Tested on three open-source LRMs across five reasoning datasets, our framework reduced token generation by 20-50\% while maintaining a comparable accuracy. 
\end{itemize}
%

\section{Related Work}

To render models less verbose and more efficient, Train and Test Time Scaling approaches have been explored \citep{qu2025surveyefficientreasoninglarge,li202512surveyreasoning,chen2025reasoningerasurveylong}. Also, recent work has explored monitoring the generation of LRMs.


\noindent \textbf{Efficient Reasoning during Inference.} Researchers have explored Inference Time Scaling technique to increase the efficiency of models \citep{qu2025surveyefficientreasoninglarge}. \emph{Model Switch} uses a router module to select small or large models for inference depending on the complexity of the problem \citep{ong2025routellmlearningroutellms}. Similarly, \emph{System Switch} looked at dynamically selecting inference settings based on the problem \citep{aytes2025sketchofthoughtefficientllmreasoning}. \emph{Length Budgeting} aims to reduce the budget allocated to the generation of answers. Works such as \citet{lee2025llmscompresschainofthoughttoken, han2025tokenbudgetawarellmreasoning, xu2025chaindraftthinkingfaster} showed that careful prompt engineering can lead to more efficient generation compared to standard inference. In addition, \citet{pu2025thoughtterminatorbenchmarkingcalibratingmitigating} demonstrated that calibration experiments can be performed to estimate the optimal number of tokens to solve particular problems. However, these techniques rely on fixed budgets or template constraints defined before inference and therefore do not adapt to the evolving reasoning traces of the model. As a result, they cannot exploit the information contained in the model's own generated reasoning during inference. In contrast, \citet{yang2025dynamicearlyexitreasoning} and \citet{wang2026entropythinkreasoningmodel} proposes dynamic early stopping criteria based on the model's confidence (DEER), or entropy (EAT), respectively. Even though these methods are dynamic based on the model's generation, they requires full or partial access to the internals, operating in a white-box (DEER), or gray-box (EAT) settings. We address this by suggesting a dynamic black-box early-exit criteria, operating only on the text generated (no use of internals or proxy models).

\noindent \textbf{Monitoring LRM generation.} We observe an emerging theme of research on monitoring LRM generation at a step level. Specifically, \citet{lee2025evaluatingstepbystepreasoningtraces} proposes a taxonomy of reasoning traces evaluators, but acknowledge that existing monitoring approaches are not adapted to complex reasoning traces. Similarly, \citet{golovneva2023roscoesuitemetricsscoring} prompts a model to generate step-by-step reasoning, and defines an error-type taxonomy to evaluate reasoning steps. Nevertheless, this method is post-hoc and focuses on measuring the quality of reasoning. Closer to our work, LCoT2Tree converts long CoT into hierarchical tree structures to analyse the reasoning patterns of LRMs \citep{jiang-etal-2025-makes}. The framework classifies thoughts into types of reasoning, but their framework is applied after the generation, and is not applied dynamically. Further, \citet{venhoff2025basemodelsknowreason} derive a taxonomy of reasoning behaviors through trained Sparse Auto-Encoders on step activations, clustering them to identify common behaviors. While their approach results in a rich taxonomy, their technique requires full-access to the model's internals. \citet{yu2026explainablechainofthoughtreasoningempirical} also note that \emph{``the high-level semantic roles of reasoning steps and their transitions remain underexplored."}. As a result, existing works often overlook the question of how to dynamically monitor LRMs reasoning during single inferences from text alone, and apply such monitoring for early-stopping purposes. 



\section{Step-Tagging module} \label{sec:step-tagging-module}

In this section, we introduce \emph{Step-Tagging}, a lightweight module capable of identifying, and tagging reasoning steps in real-time during inference.

\textbf{Step-by-step generation.} To enable fine-grained monitoring of LRMs, we decompose the reasoning traces into step sequences. Let $y = y_{1:n} \in V^n$ be the output token sequence generated by the model over the vocabulary $V$. We segmented LRM's outputs as follows:
\begin{equation}\label{eq:step-equation}
S = \{s_1, \ldots, s_i, \ldots, s_{T}\}, \text{ such as } s_i = y_{r_{i-1}:r_{i}}
\end{equation}
where $r_{i}$ corresponds to the delimiters of the step segments. In this paper, we segmented the reasoning traces based on the delimiter $\alpha = \texttt{"\textbackslash n\textbackslash n"}$ \citep{zhang2025reasoningmodelsknowtheyre}. 

\textbf{Objective.} Our definition of a reasoning step enables users to segment reasoning steps within model outputs. However, this definition alone does not allow the user to annotate the segmented steps with reasoning types. This annotation would enable users to track logical transitions within model outputs. To do this, we must first define a tag dictionary $\mathcal{T}_{\text{tags}}$ (i.e., a label space of reasoning step tags) that covers the types of reasoning steps generated by models. Essentially, given a sequence of reasoning steps \( S \), we wish to label each step \( s_i \) with a tag \( \tau_i \in \mathcal{\mathcal{T}_{\text{tags}}} \). Formally, we are looking to construct a step-tagging function $\phi$ such as:
\begin{equation}\label{eq:step_tagging}
    \forall \hspace{0.1cm} i \in [1,T], \phi(s_i) = \tau_i
\end{equation}
where $s_i \in S$ is a reasoning step from the full output sequence $y$, $\phi$ is the step-tagging function, and $\tau_i \in \mathcal{T}_{tags}$ is the reasoning tag associated to the step $s_i$.

\textbf{Taxonomy of the type of steps.} To enable fine-grained monitoring of reasoning behavior, we need to know the different types of reasoning steps that are typically generated by LRMs (i.e., we need to define $\mathcal{T}_{\text{tags}}$). To do so, we created a taxonomy based on the outputs of both DeepSeek-R1-Distill-Llama-8B \citep{deepseekai2025deepseekr1incentivizingreasoningcapability} and QwQ-32B \citep{qwq32b} models.

\begin{figure}[h]
    \centering
    \includegraphics[width=1.0\linewidth]{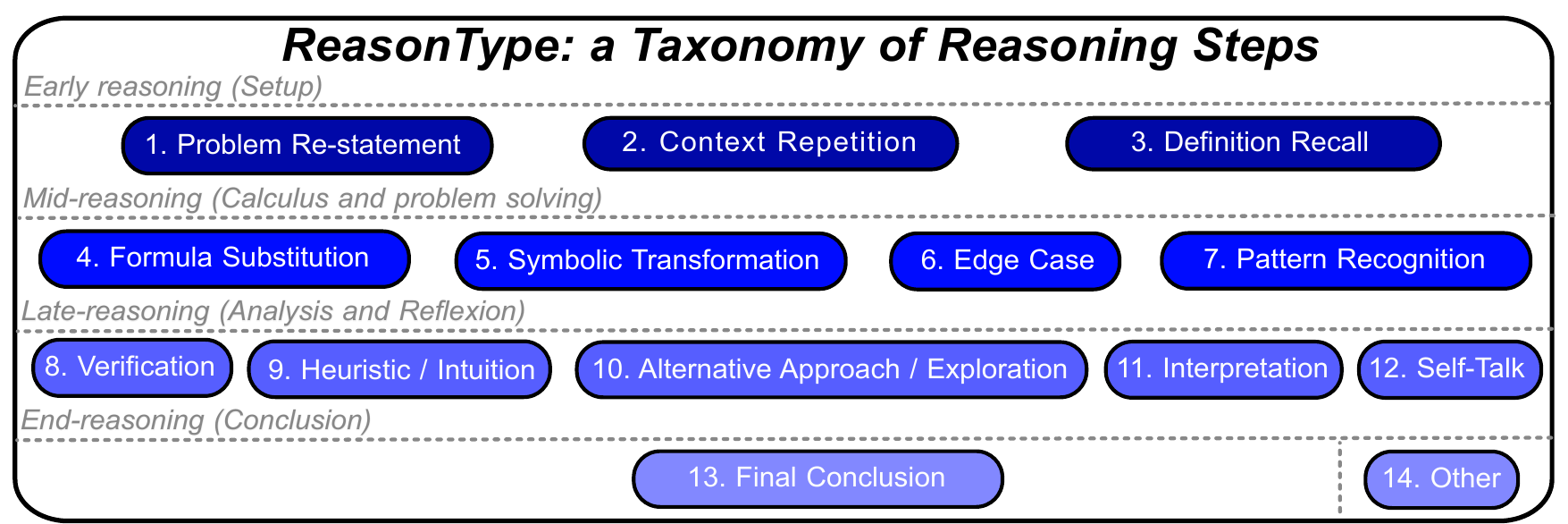}
    \vspace{-0.3cm}
    \caption{ReasonType - A taxonomy of reasoning step types as per \texttt{gpt-4o-mini}}
    \label{fig:taxonomy-step-type}
    \vspace{-0.6cm}
\end{figure}

Inspired by prior work on model behavior analysis \citep{galichin2025icoveredbaseshere, kuznetsov2025featurelevelinsightsartificialtext}, we first created a prompt to identify distinct types of reasoning steps in the traces. We then sampled $100$ reasoning traces from the MATH500 train dataset (covering $20$ samples per difficulty level for each model) and using our prompt submitted the traces to \texttt{GPT-4o-mini} \citep{openai2024gpt4technicalreport}. The prompt resulted in a series of different step-types. We merged overlapping categories, to construct a taxonomy that reflects the temporal and reasoning progression of model's traces. We refer to this taxonomy as \emph{ReasonType} (Figure \ref{fig:taxonomy-step-type}) encompassing $13$ categories, including early-stage behaviors such as \emph{Problem Re-statement}, or later reasoning stages like \emph{Verification} and \emph{Exploration}. We detailed our construction and validation process in Appendix \ref{app:constr-valid-taxonomy}. 

\textbf{Early-stopping criteria.} In the following section, we see that LRMs tend to generate the answer early in the output sequence, with step-types following an ordered pattern. Based on this observation, the central challenge that we address is to determine when to stop the generation of LRMs based on step tags, creating an interpretable stopping criterion. Assuming that our Step-Tagging framework can effectively monitor the steps (Equation \ref{eq:step_tagging}), we can define a constraint on the frequency of a given step type. Each constraint operates online, over a running sequence of reasoning steps $S_{\text{running}} = \{s_1, \dots, s_j\}$, where each step $s_i$ is associated with a tag $\tau_i \in \mathcal{T}$. We define the constraint $c_{\tau^*}$ as:
\vspace{-0.3cm}
\begin{equation}\label{eq:early_stopping_step_tag}
    c_{\tau^*}(S_{\text{running}}, \delta) = \mathbf{1}[f_{\text{freq}}(S_{\text{running}}, \tau^*) \leq \delta] 
\end{equation}
\vspace{-0.4cm}
\begin{equation*}
    \text{ with } f_{\text{freq}}(S_{\text{running}}, \tau^*) = \sum_{i=1}^j \mathbf{1}[\tau_i = \tau^*]
\end{equation*}

\vspace{-0.5cm}

where $c_{\tau^*}(S_{\text{running}}, \delta)$ is the constraint on the tag type $\tau^*$ over the step-sequence $S_{\text{running}}$ being generated, given the threshold $\delta$. $f_{\text{freq}}(S_{\text{running}}, \tau^*)$ is the occurrence of the type-step $\tau^*$ over the running sequence $S_{\text{running}}$. While the constraint $c_{\tau^*}$ is satisfied, the generation continues. If the constraint is violated, the generation stops (see Algorithm \ref{alg:st_es_algorithm} in Appendix \ref{sec:appendix-es_algorithm} for more implementation details).

To facilitate the evaluation of early-exit answers, we prompted the models right after early-stopping, and allowed an additional budget of $100$ tokens. We borrowed this approach from \citet{muennighoff2025s1simpletesttimescaling}, who showed that this intervention helped the model to provide its current best answer.

\section{Experimental Setting} \label{sec:experimental-section}

\begin{figure*}[ht]
    \centering
    \begin{subfigure}{0.33\linewidth}
        \centering
        \includegraphics[width=\linewidth]{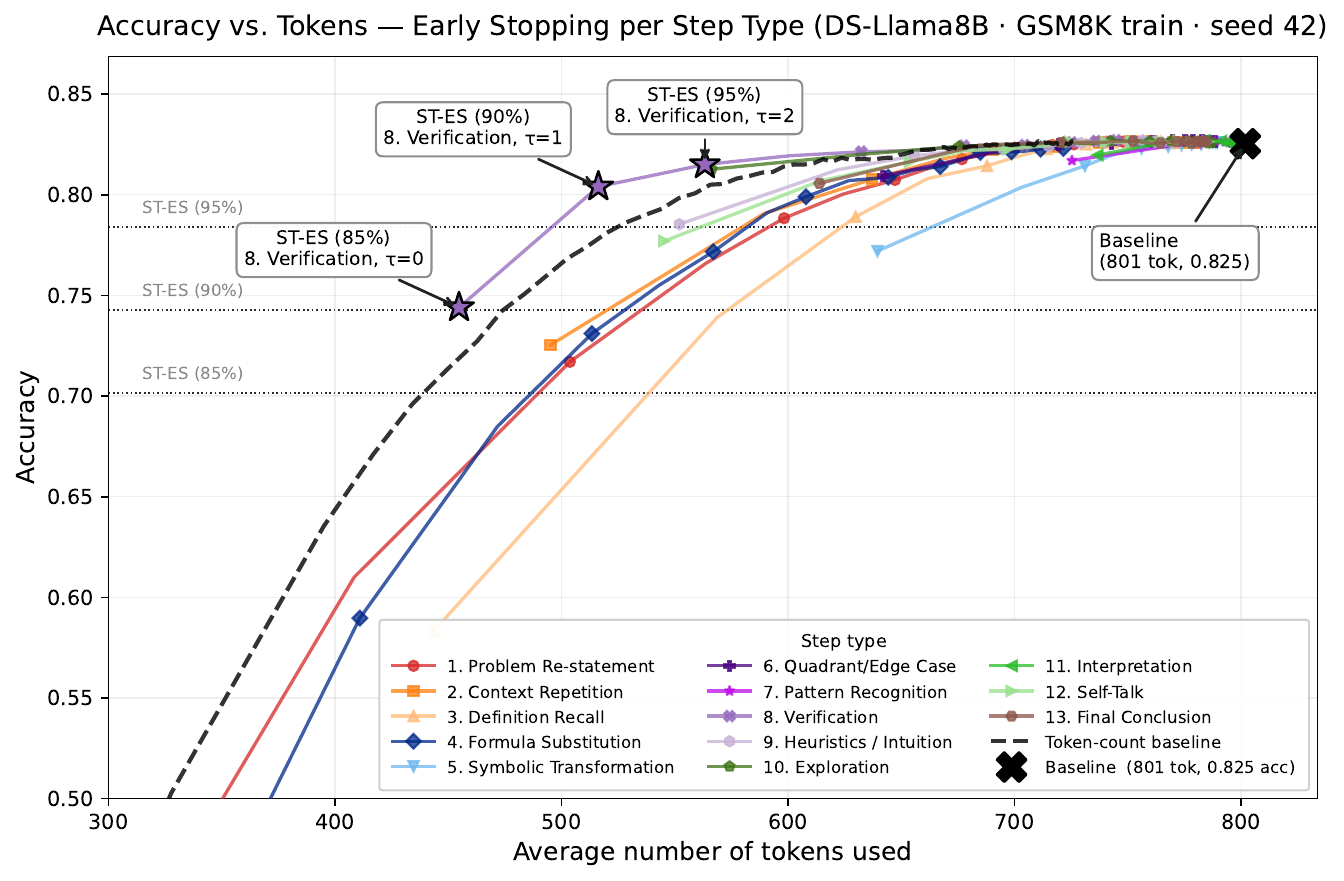}
        \caption{DS-Llama8B on GSM8K}
        \label{fig:DS8B-GSM8K} 
    \end{subfigure}
    \hfill
    \begin{subfigure}{0.33\linewidth}
        \centering
        \includegraphics[width=\linewidth]{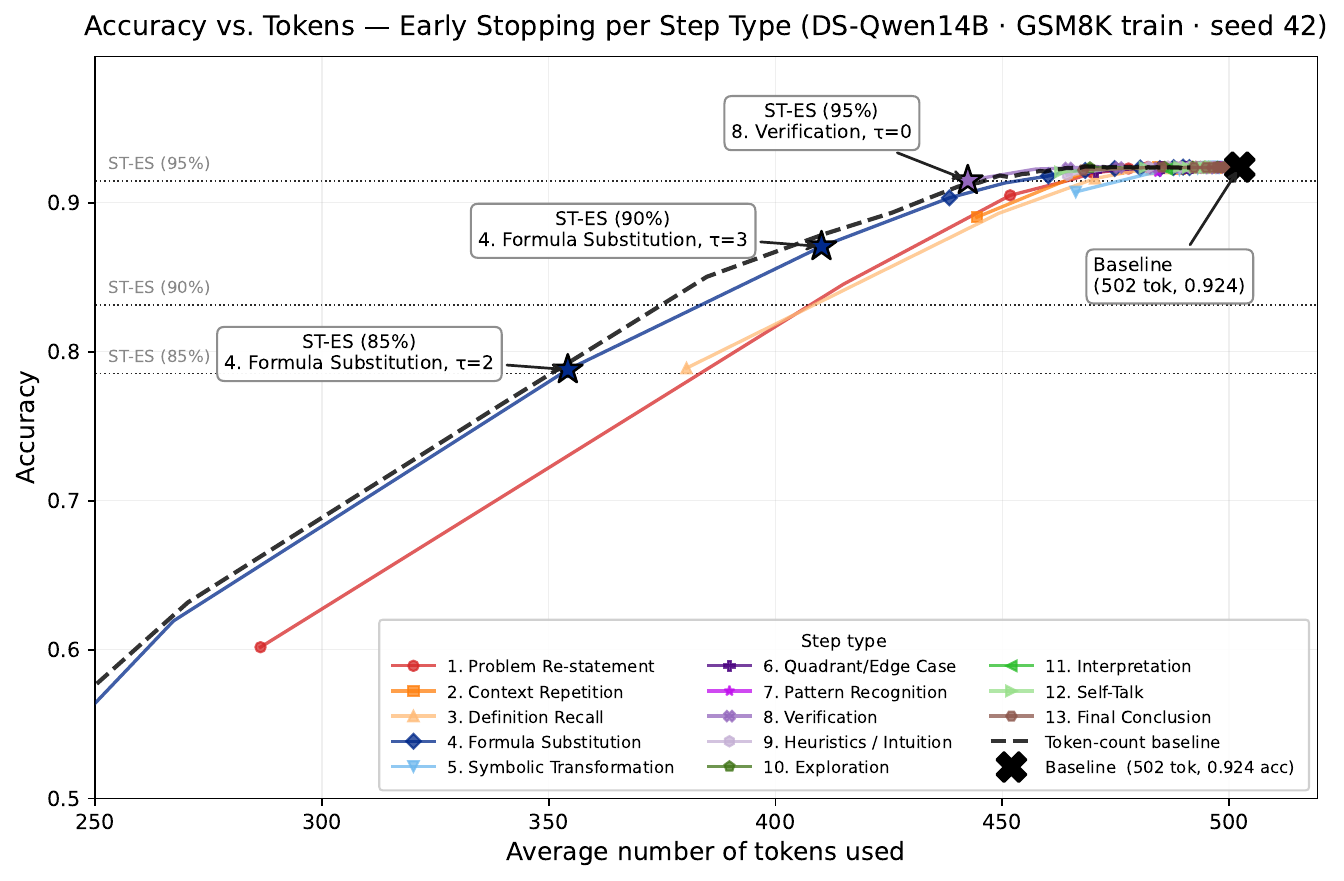}
        \caption{DS-Qwen14B on GSM8K}
        \label{fig:DS14B-GSM8K}
    \end{subfigure}
    \hfill
    \begin{subfigure}{0.33\linewidth}
        \centering
        \includegraphics[width=\linewidth]{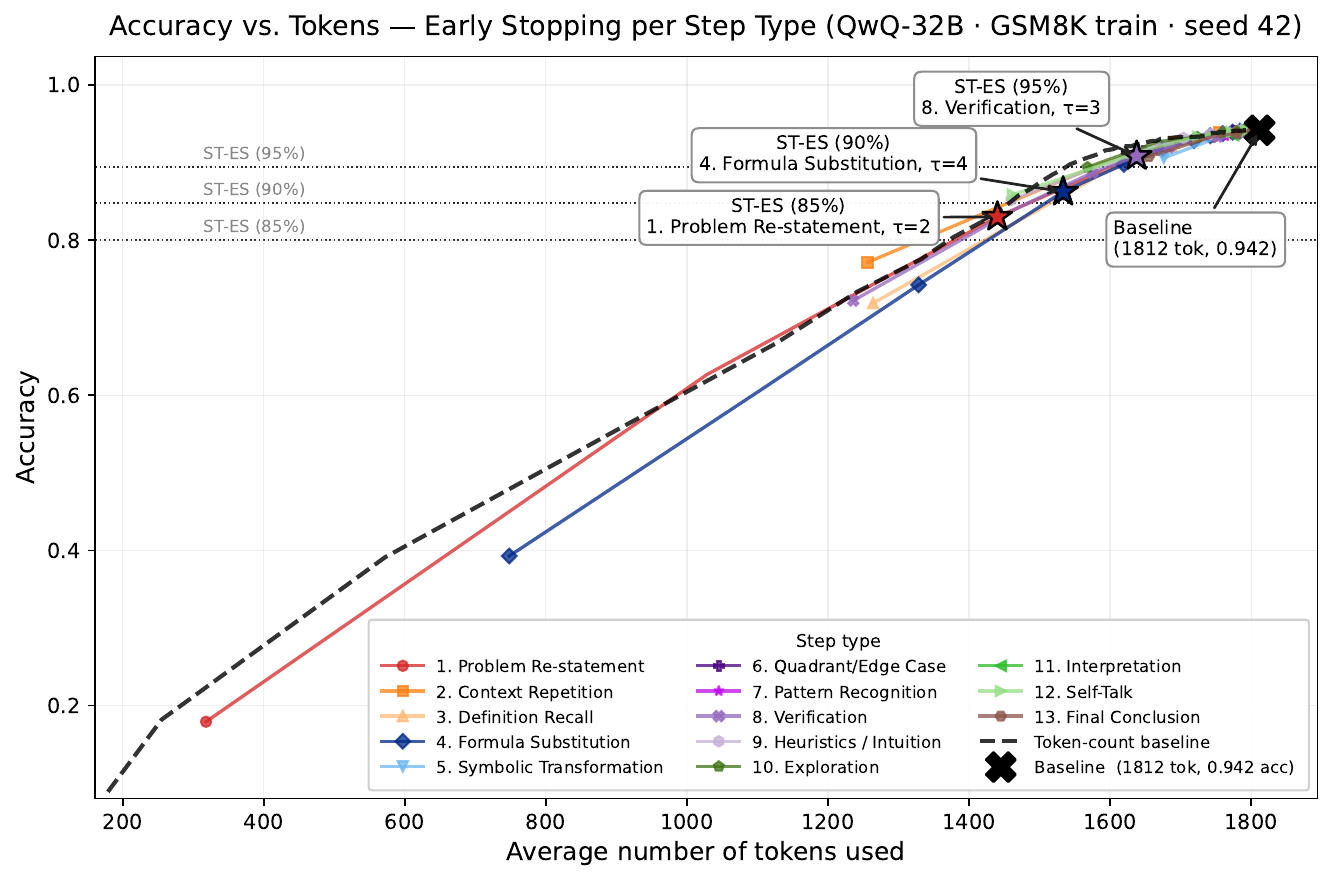}
        \caption{QwQ-32B on GSM8K}
        \label{fig:QwQ32B-GSM8K}
    \end{subfigure}
    \begin{subfigure}{0.33\linewidth}
        \centering
        \includegraphics[width=\linewidth]{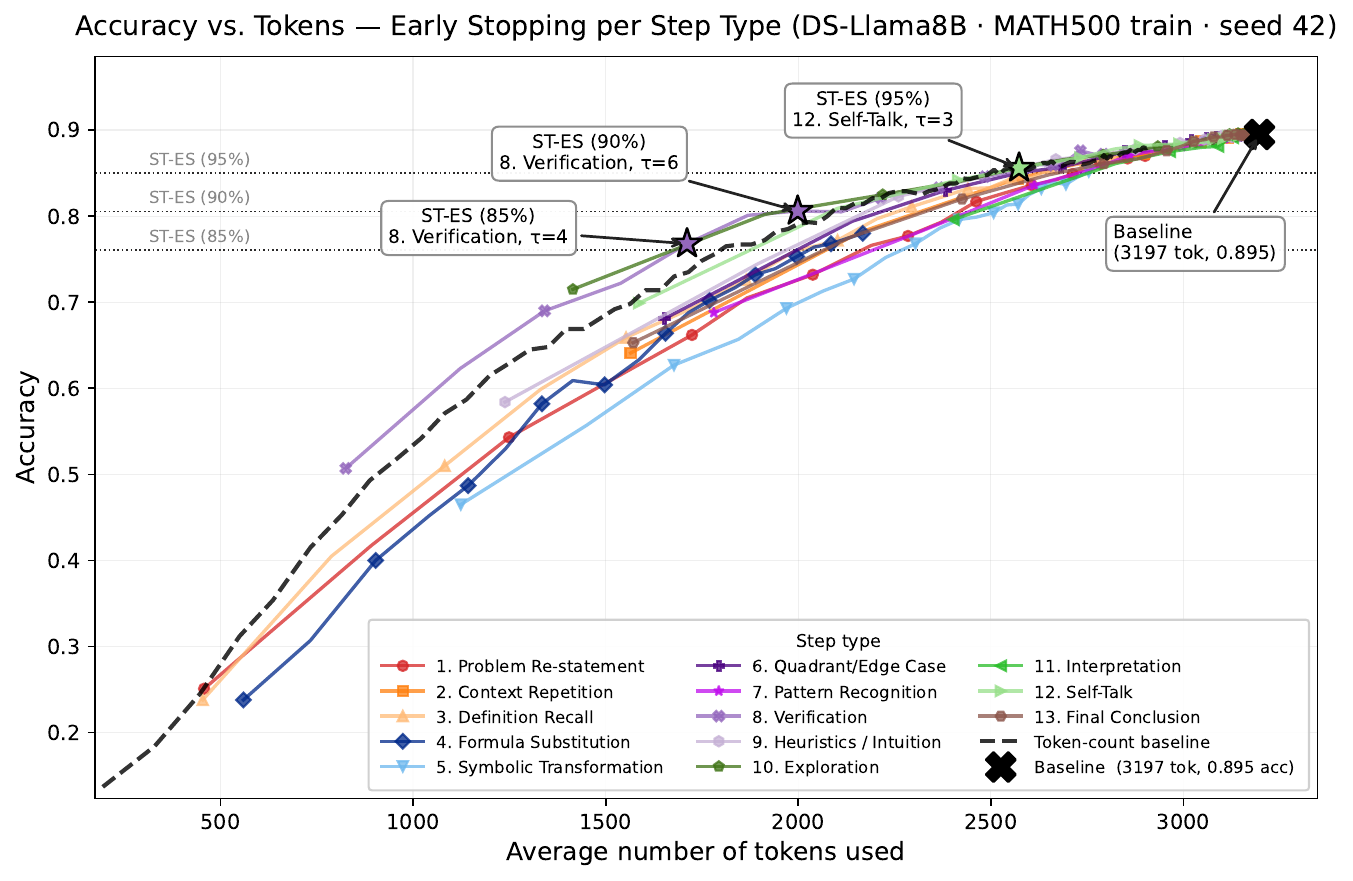}
        \caption{DS-Llama8B on MATH500}
        \label{fig:DS8B-MATH500} 
    \end{subfigure}
    \hfill
    \begin{subfigure}{0.33\linewidth}
        \centering
        \includegraphics[width=\linewidth]{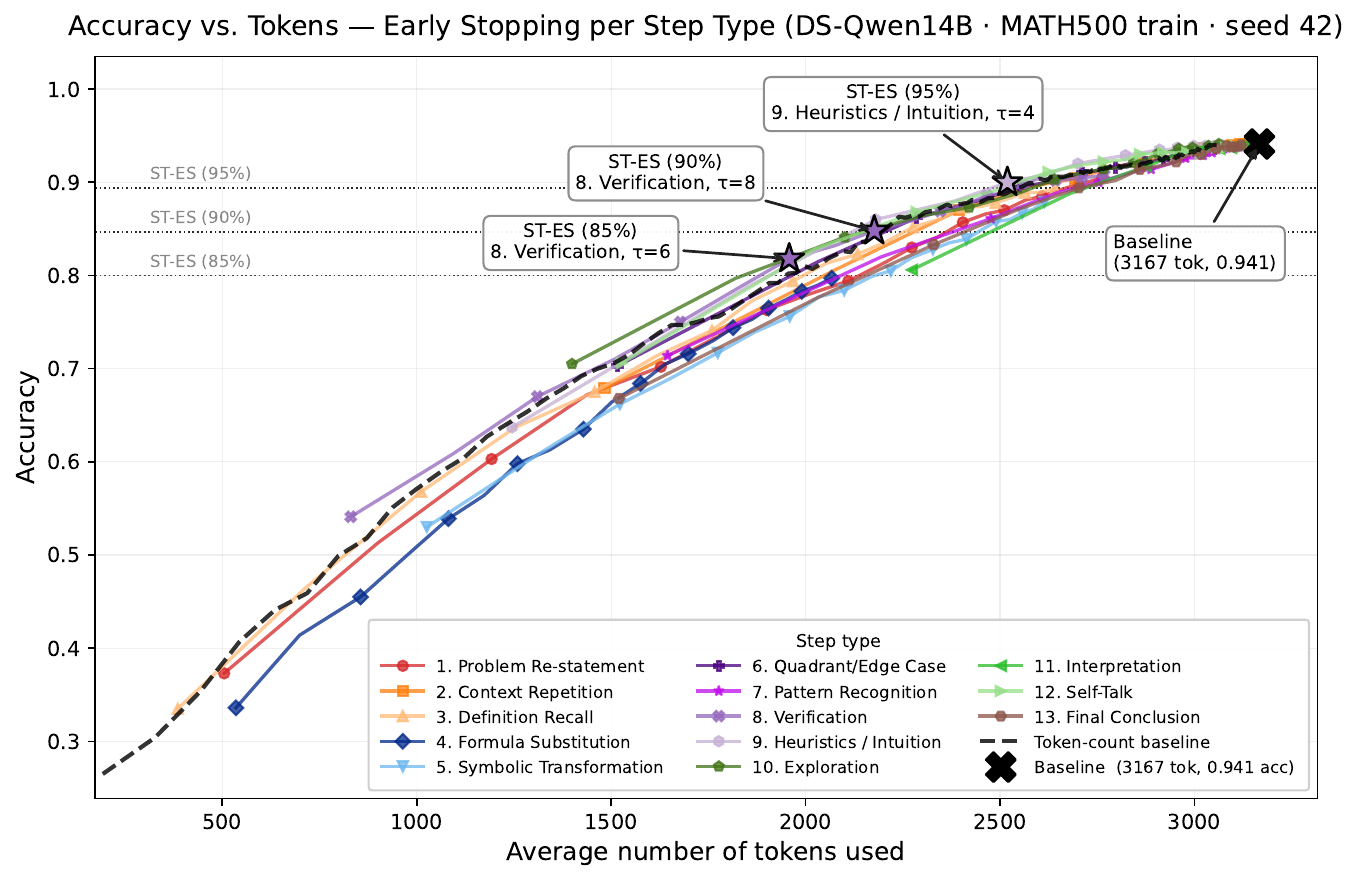}
        \caption{DS-Qwen14B on MATH500}
        \label{fig:DS14B-MATH500}
    \end{subfigure}
    \hfill
    \begin{subfigure}{0.33\linewidth}
        \centering
        \includegraphics[width=0.99\linewidth]{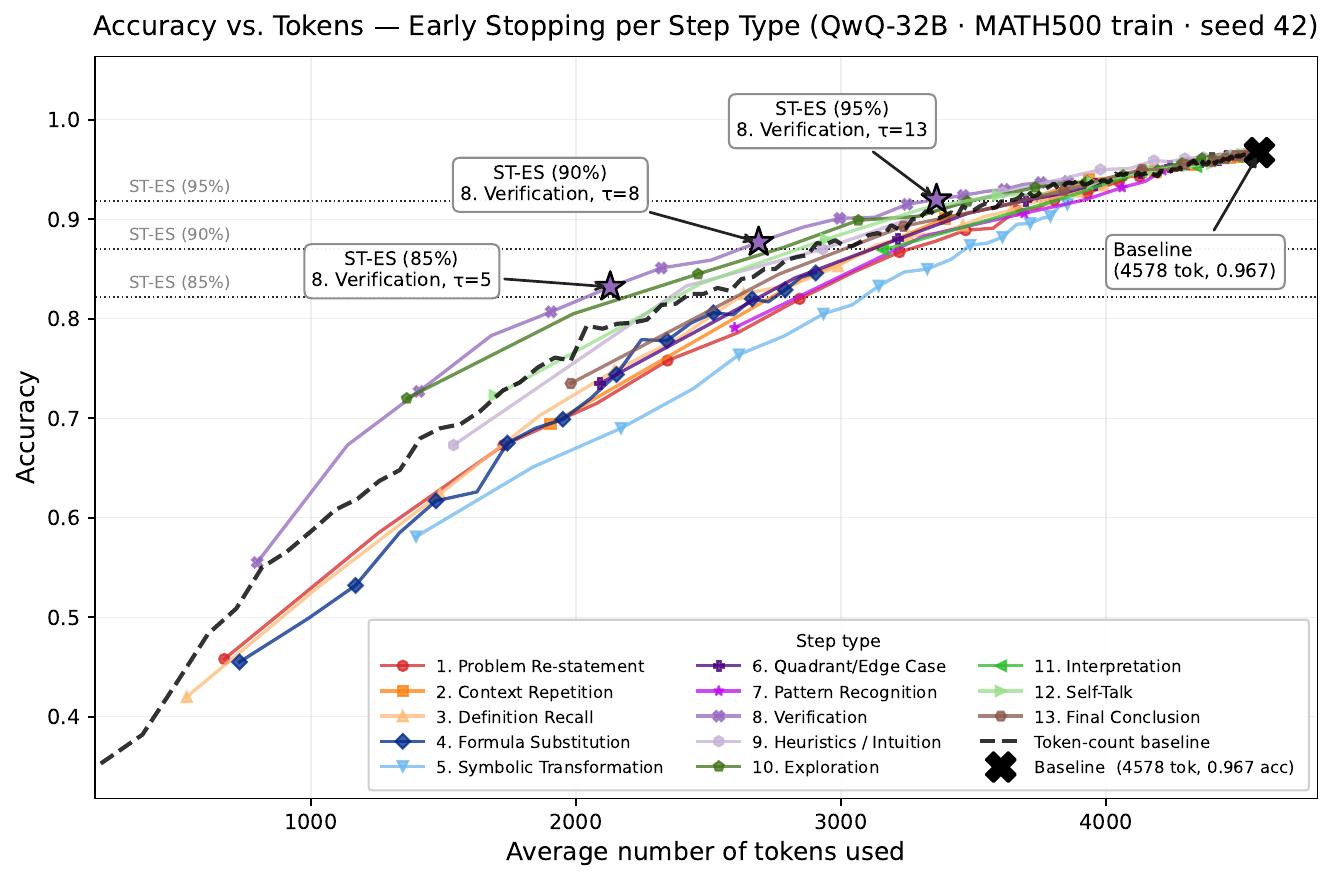}
        \caption{QwQ-32B on MATH500}
        \label{fig:QwQ32B-MATH500}
    \end{subfigure}
    \\
    \begin{subfigure}{0.33\linewidth}
        \centering
        \includegraphics[width=\linewidth]{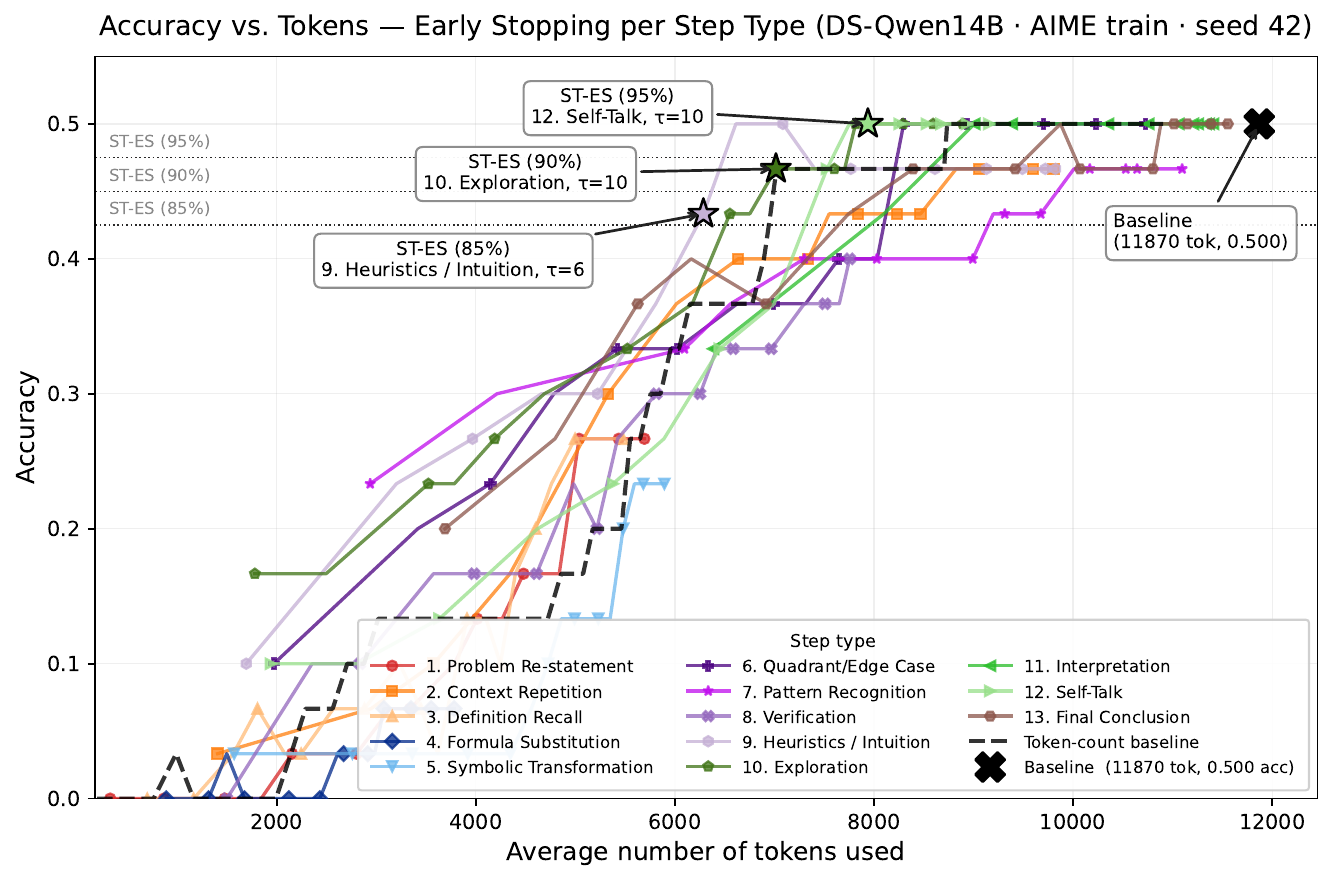}
        \caption{\small DS-Qwen14B on AIME}
        \label{fig:DS14B-AIME} 
    \end{subfigure}
    \hfill
    \begin{subfigure}{0.33\linewidth}
        \centering
        \includegraphics[width=\linewidth]{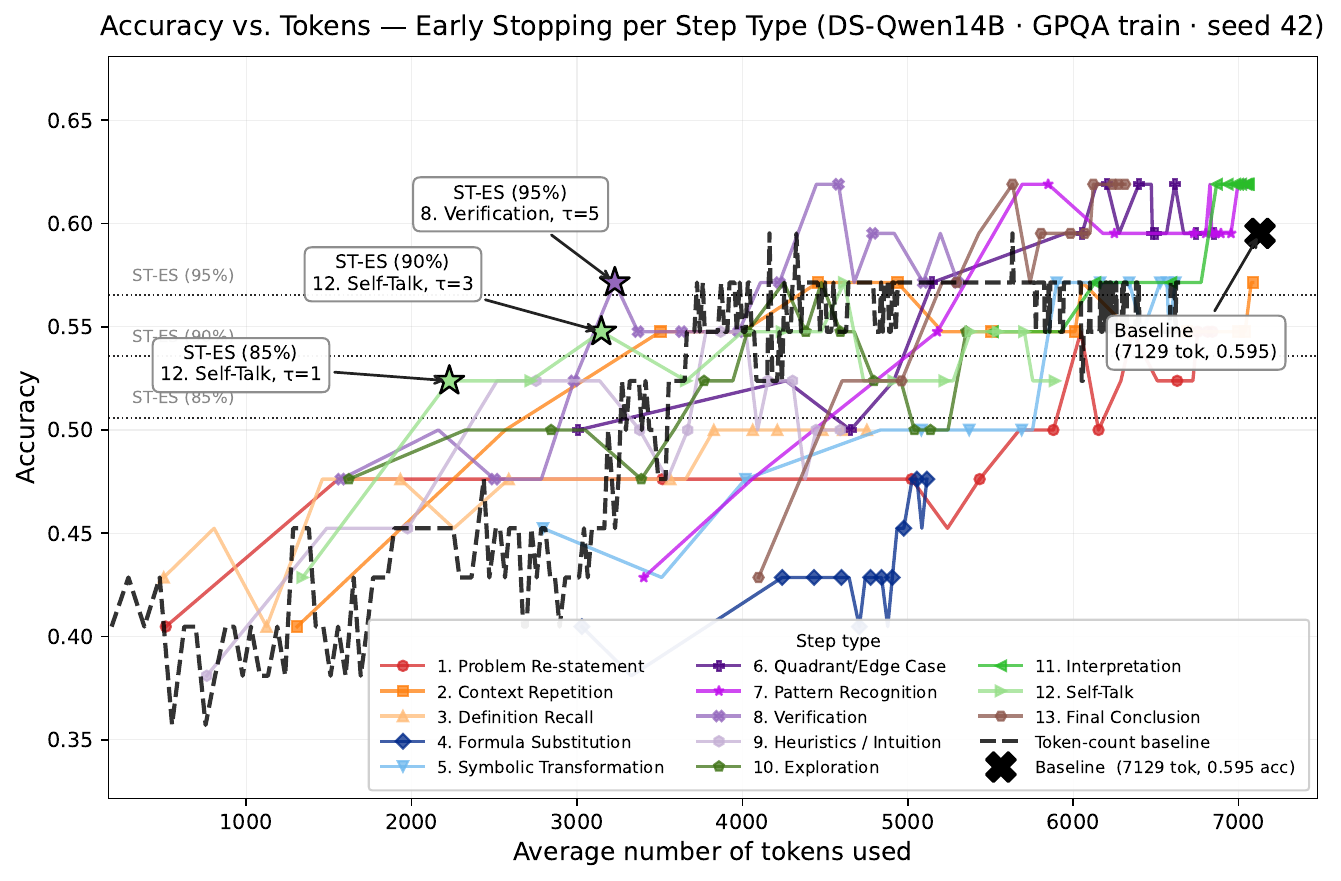}
        \caption{\small DS-Qwen14B on GPQA}
        \label{fig:DS14B-GPQA}
    \end{subfigure}
    \hfill
    \begin{subfigure}{0.33\linewidth}
        \centering
        \includegraphics[width=\linewidth]{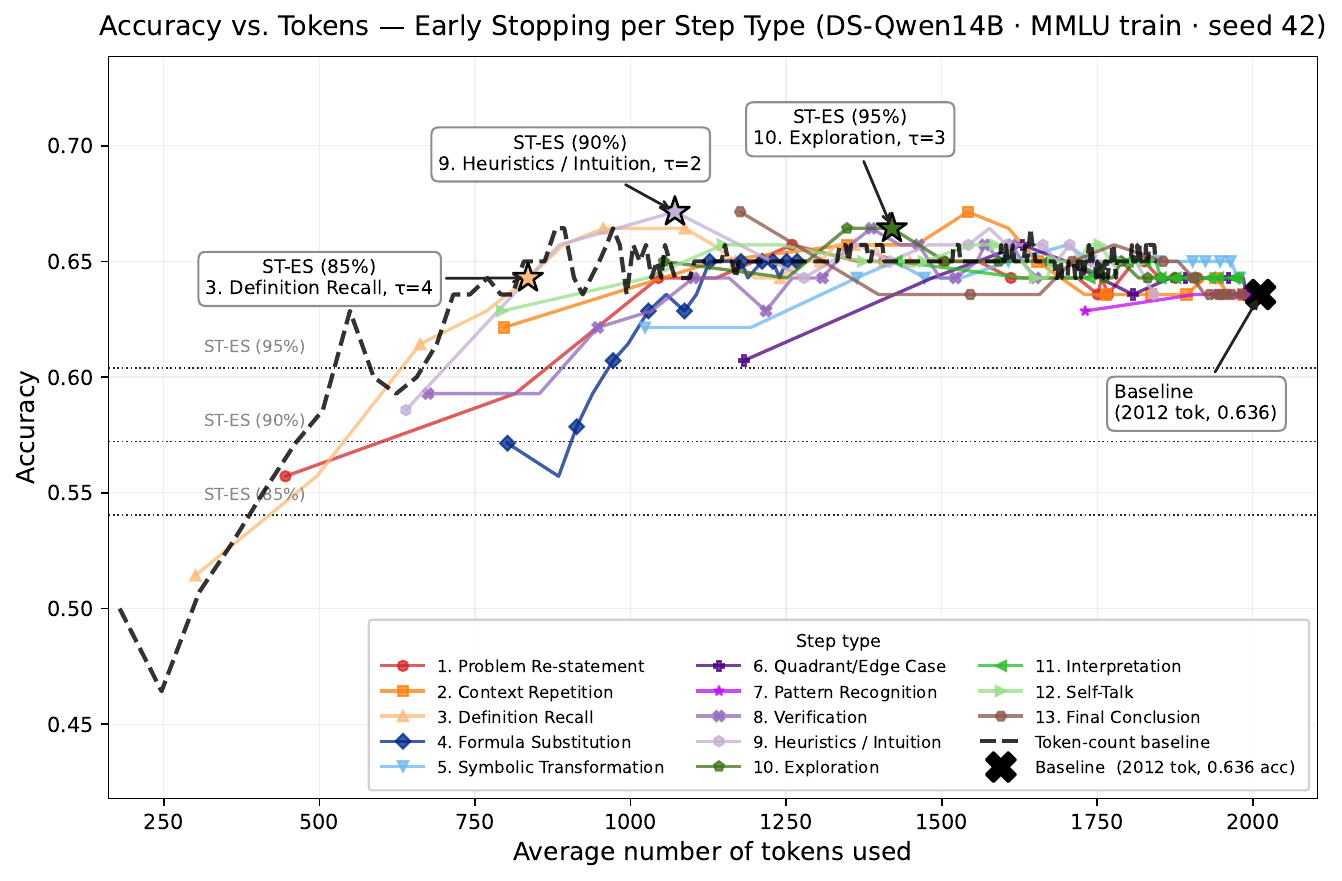}
        \caption{\small DS-Qwen14B on MMLU-Pro}
        \label{fig:DS14B-MMLU}
    \end{subfigure}
    \vspace{-0.3cm}
    \caption{\small Number of Tokens vs. Pass@$1$ - ST-ES criteria on train datasets for $\delta \in [0,20]$ - Each curve represent an ST-ES criteria with annotated step-types by \texttt{GPT-4o-mini}. We note that the efficiency curves differs depending on the step type. ST-ES criteria based on verification or self-reflection steps tends to offer the best accuracy trade-offs, almost systematically beating the token-count baseline.} 
    \label{fig:st-es-calibration}
    \vspace{-0.5cm}
\end{figure*}

Our paper contains two objectives. First, our goal is to compare the contribution of each step-types to the efficiency of LRMs, and identify the most reliable early-stopping configurations. Second, we assess whether the ST-ES framework can be applied during deployment. In this section, we will first motivate our selection of models, datasets, and metrics,  followed by our baseline implementation.

\textbf{Datasets and metrics.} To assess our approach, we selected five state-of-the-art reasoning datasets. First, we evaluated our models on three mathematical datasets, namely: \emph{GSM8K}, \emph{MATH500}, and \emph{AIME}. Second, we evaluated the applicability of our framework on different domains, with \emph{GPQA-Diamond} and \emph{MMLU-Pro}. We evaluated the correctness of the answers using Math-Verify (mathematic tasks) or MCQ-Prompting (knowledge and reasoning benchmarks). Our analysis in Section \ref{sec:influence-step-type} and the training of our step-tagger (Section \ref{sec:monitoring-step-tagging}) is conducted on the \emph{training splits}. We then evaluated ST-ES in Section \ref{sec:STES-performance} on the held-out \emph{testing splits}. Table \ref{tab:reasoning-datasets} in Appendix \ref{sec:appendix-experimental-design} presents the datasets that we selected. 

\textbf{Model selection.} To apply our framework a user must have access to the fine-grained reasoning traces of LRMs. Open-source models like DeepSeek-R1 and QwQ consistently provide reasoning traces. For this reason, we focus our analysis exclusively on \texttt{DeepSeek-R1-Distill-Llama-8B}, \texttt{DeepSeek-R1-Distill-Qwen-14B} and \texttt{QwQ-32B}, which offer the granularity needed to monitor the reasoning process. This choice is motivated by their variety in term of size and performance, and diversity in providers. 

\noindent \textbf{Inference setting.} For the purposes of our experiments, we performed standard inference and applied our \emph{Step-Tagging} and \emph{Early-Stopping} algorithms \emph{offline}. To ensure the robustness and reproducibility of our approach, we generated outputs using fixed random seeds with deterministic decoding. We used five seeds for GSM8K and MATH500 across all three model sizes (8B, 14B, 32B), and we used a single seed on AIME, GPQA and MMLU on the 14B model.

\noindent \textbf{Baselines.} To assess the effectiveness of our early-stopping approach, we define two baselines. First, we introduce a \emph{Token-count} baselines, which early-stops the generation for a given count of step (regardless of their step-type) \cite{pu2025thoughtterminatorbenchmarkingcalibratingmitigating}. Second, we set-up \emph{Prompt-guided} baselines, which explicitly instruct the models to not generate verbose output, or over-verification steps \citep{lee2025llmscompresschainofthoughttoken}. We implemented user-prompt and system-prompt variants, with Zero-Shot and Few-Shot prompts that aim to reduce the reasoning computation while retaining accuracy. We selected 4 variants, namely: zero-shot user and system prompt, and few-shot system prompt with $1$ and $3$ examples: $\mathcal{P}^{(0)}_{\text{user}}$, $\mathcal{P}^{(0)}_{\text{system}}$, $\mathcal{P}^{(1)}_{\text{system}}$, $\mathcal{P}^{(3)}_{\text{system}}$, respectively. The prompts used to establish these baselines are listed in Appendix \ref{sec:appendix-experimental-design}.

\section{Influence of step-types on LRM's efficiency} \label{sec:influence-step-type}

Our first objective is to determine whether tracking the type of reasoning steps is useful to early-stop the generation of LRMs. To do so, we first inferred models on the train split of the dataset, annotated their reasoning steps using \texttt{GPT-4o-mini}, and applied ST-ES. Figure \ref{fig:st-es-calibration} presents the number of tokens vs. accuracy of every tag-type with values of threshold ranging from $0$ to $20$, for each configurations. Each colored curve represent our ST-ES criterion applied to a specific step-type. The doted black curve represent the Token-Count baseline (for $\delta \in [1,100-500]$).

\textbf{Step-type, a useful signal.} First, we observe that the ST-ES curves systematically matches or outperforms the Token-Count baselines for all configurations. For instance, in Figures \ref{fig:DS8B-MATH500}-\ref{fig:QwQ32B-MATH500} (MATH500), at least one curve stands above the black doted line. It means that for equivalent token-count, ST-ES achieves higher accuracy. This observation holds across the three models and other datasets.

\textbf{Step-types of ST-ES.} Second, each step-type offer a distinct accuracy/token-count trade-offs. This is particularly visible on GSM8K (Figure \ref{fig:DS8B-GSM8K}), and MATH500 (Figures \ref{fig:DS8B-MATH500}-\ref{fig:QwQ32B-MATH500}), where late reasoning step-types such as \emph{Verification} or \emph{Self-Talk} stand above early step-types (e.g. \emph{Problem Re-Statement}). For instance, in Figure \ref{fig:DS14B-MATH500}, \emph{Verification} with $\tau = 5$ reaches 85\% of the accuracy with only 50\% of the token-count. Results are less pronounced on AIME and GPQA (Figures \ref{fig:DS14B-AIME} and \ref{fig:DS14B-GPQA}), where curves are noisier - likely due to a lower number of sample and higher token-count relatively to other datasets, making the analysis less stable. \\
Further, we selected three best inference configuration corresponding to expected accuracy drop (namely 85, 90 and 95\%), and evaluate them on test datasets.

\vspace{-0.15cm}

\section{Monitoring LRMs using Step-Tagger} \label{sec:monitoring-step-tagging}

\vspace{-0.15cm}

To validate our approach, we first train sentence classifiers for efficient step-type monitoring, and then evaluate our criteria on test datasets.

\begin{figure*}[t]
    \centering
    \begin{subfigure}{0.49\linewidth}
        \includegraphics[width=1\linewidth]{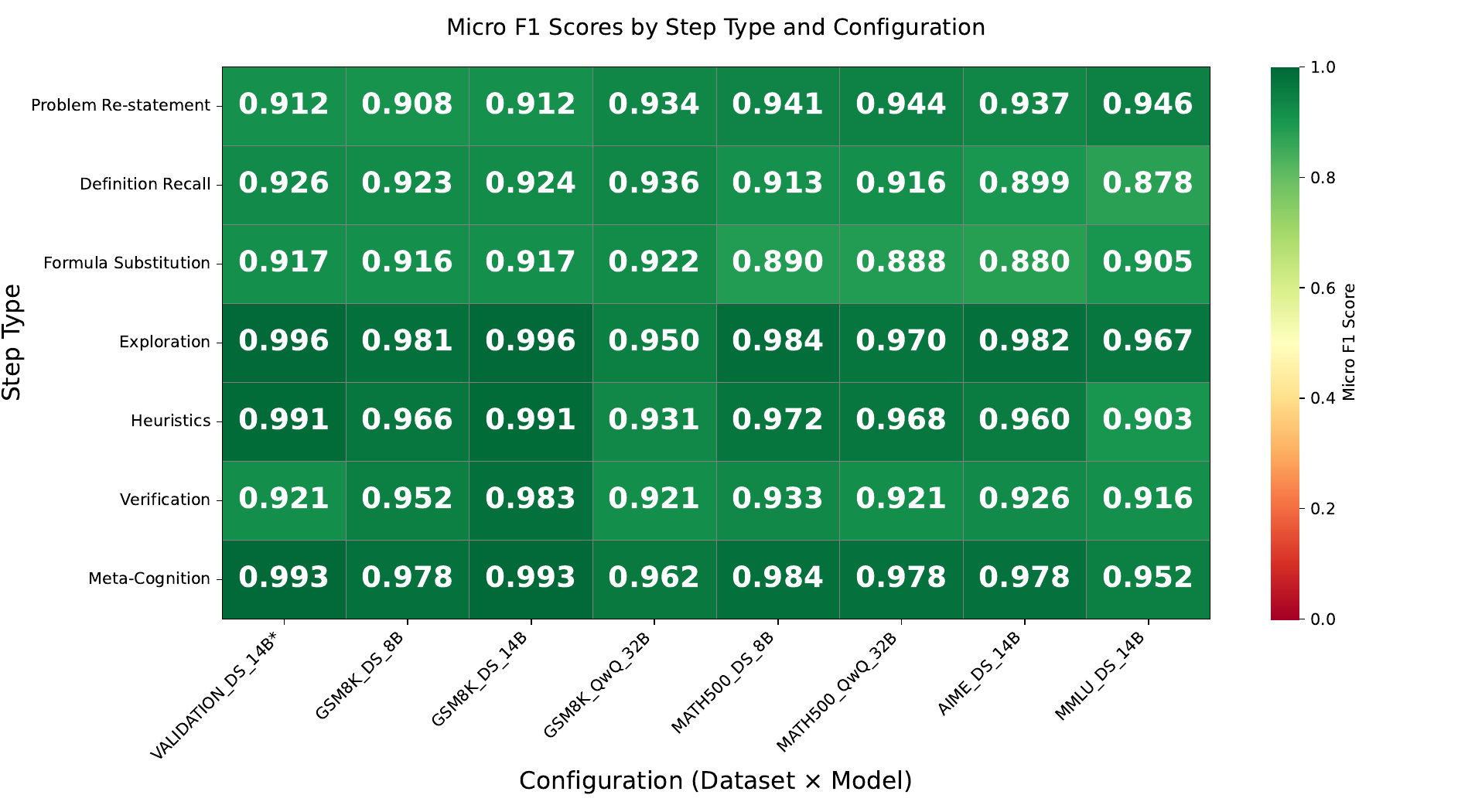}
        \caption{Micro-F1}
        \label{fig:heatmap-microf1}
    \end{subfigure}
    \hfill
    \begin{subfigure}{0.47\linewidth}
        \includegraphics[width=1\linewidth]{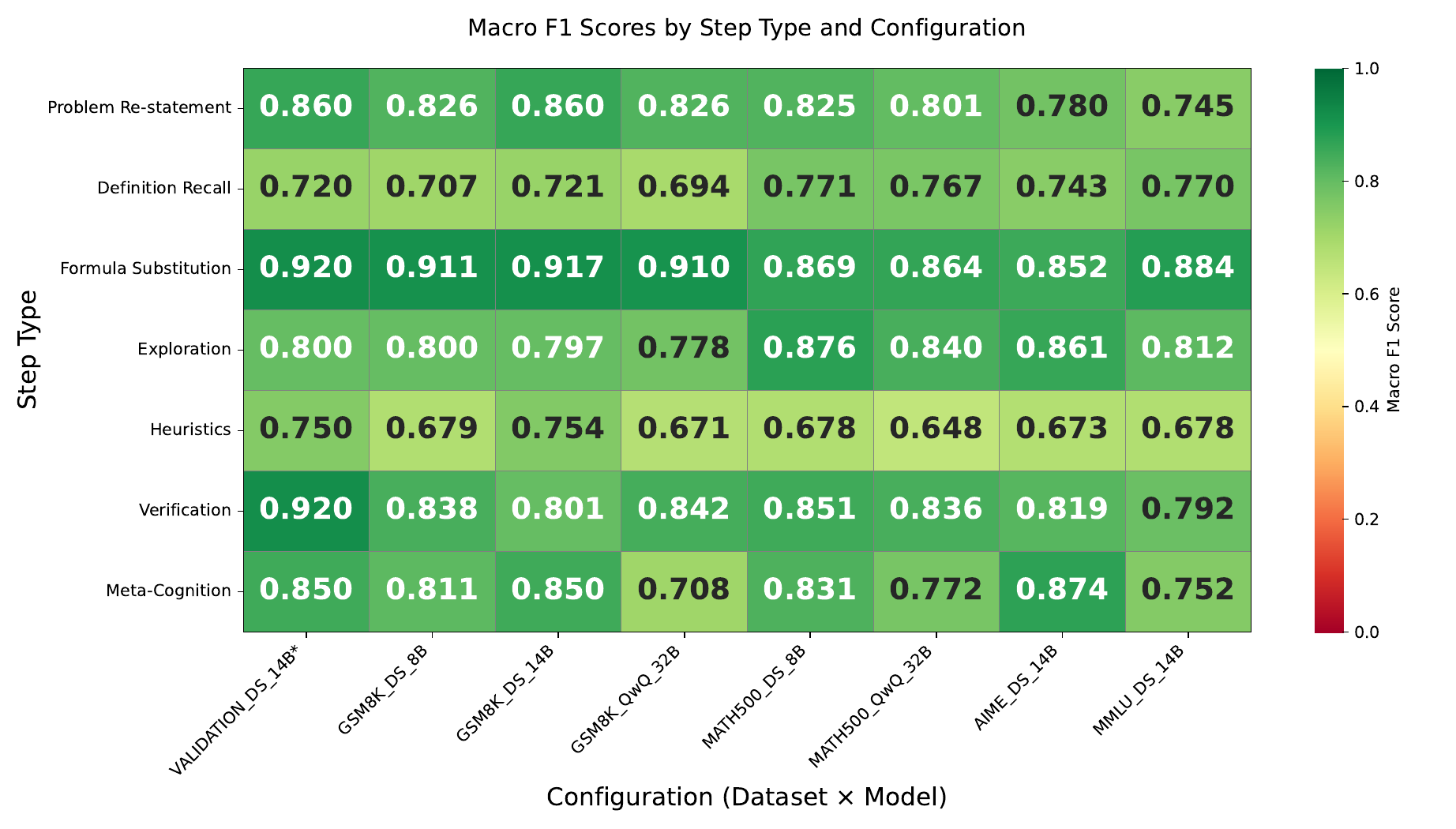}
        \caption{Macro-F1}
        \label{fig:heatmap-macrof1}
    \end{subfigure}
    \caption{Step-Tagger performance - Seed $42$ - Figures \ref{fig:heatmap-microf1} and \ref{fig:heatmap-macrof1} present the Micro and Macro F1 score of the binary BERT classifiers across selected step-types and model/dataset configurations. The first column of each heatmaps report the performance on the held-out validation set (20\% of the training set), composed of steps generated by DS-14B on MATH500 and GPQA train.} 
    \label{fig:step_tagger_performance}
\end{figure*}

\noindent \textbf{Training data generation.} Given that our reasoning step taxonomy was created using \texttt{GPT-4o-mini} \citep{openai2024gpt4technicalreport} the most direct way to label a reasoning trace would be to use \texttt{GPT-4o-mini}. However, this annotation is costly, each step requiring more than a second to be annotated (see Table \ref{tab:annotation-steps-data} in Appendix \ref{sec:appendix-labeling-process}). Consequently, instead, we used \texttt{GPT-4o-mini} to label a dataset of reasoning traces with the labels from the taxonomy that we use to train lighter weight reasoning step classifiers.  We constructed training datasets by running each LRMs on training samples of each datasets (with a seed of $42$). For each step $s_i$ in generated outputs, we prompted GPT-4o-mini to assign a tag $\tau_i$. Appendix \ref{sec:appendix-reliability-annotation} validates the reliability of GPT-4o-mini to annotate the reasoning steps. 

\noindent \textbf{Sentence classifiers.} We selected the \texttt{bert-base-uncased} sentence classifier \citep{devlin2019bertpretrainingdeepbidirectional} to construct our Step-Tagging framework, including a single hidden layer. Given the large and fine-grained nature of our taxonomy ($13$ distinct step types), training a multi-class classifier is challenging due to significant class imbalance. To address this, we trained separate binary classifiers for each step-type. This approach notably improved detection accuracy across low-frequency categories, and fits our definition of early-stopping constraint: one step-type per early-stopping criteria. To reduce the training cost, and evaluate the generalization of our apporach, we train our lightweight modules on MATH500 and GPQA train split obtained on DS-Qwen14B, and evaluate their performance on other configurations. The training details are presented in Appendix \ref{sec:appendix-training-details}.

\textbf{Classification metrics.} To evaluate the performance of our classifiers, we computed the Macro-F1 and Micro-F1 on the test datasets. While the \emph{Macro-F1} helps to identify the classifier's ability to detect rare classes, the \emph{Micro-F1} offers a more global view on the step detector's performance across all steps.

\noindent \textbf{Performance of step monitoring.} Next, our objective is to demonstrate that our approach can effectively identify the step-types in reasoning traces. Figure \ref{fig:step_tagger_performance} present the performance of the binary step-classifiers on the seven selected step-types for every configurations. We observe that the Micro-F1 is generally high across most steps for all models across all datasets - $0.88$ to $0.99$, which demonstrates that the classifiers are good at detecting step-tags. Moreover, we also reported the macro-F1 score since the distribution of step-types is highly imbalanced.

We observe lower scores, notably for \emph{Heuristics} with $0.69$ Macro-F1 on average (\emph{Heuristics} is a rare step type, representing less than $1.5$\% of the labels, and so we attribute this relatively low score to label imbalance). However, the scores remain relatively high, particularly for \emph{Verification} and \emph{Formula Substitution}. Importantly, we note that results on unseen configurations (other dataset and model configuration, i.e. DS-Llama8B, QwQ-32B, GSM8K, AIME and MMLU) remains satifying. It means that our step-taggers are transferable, and are still able to accurately identifying a given step-types for other models and tasks.

We interpret the strong performance of the classifiers as validating our reasoning step taxonomy in the sense that it indicates that the step types are distinct (i.e., they reflect types with separable properties). 




\section{Step-Tagging Early-stopping (ST-ES)} \label{sec:STES-performance}

\begin{figure*}[ht]
    \centering
    \begin{subfigure}{0.33\linewidth}
        \centering
        \includegraphics[width=\linewidth]{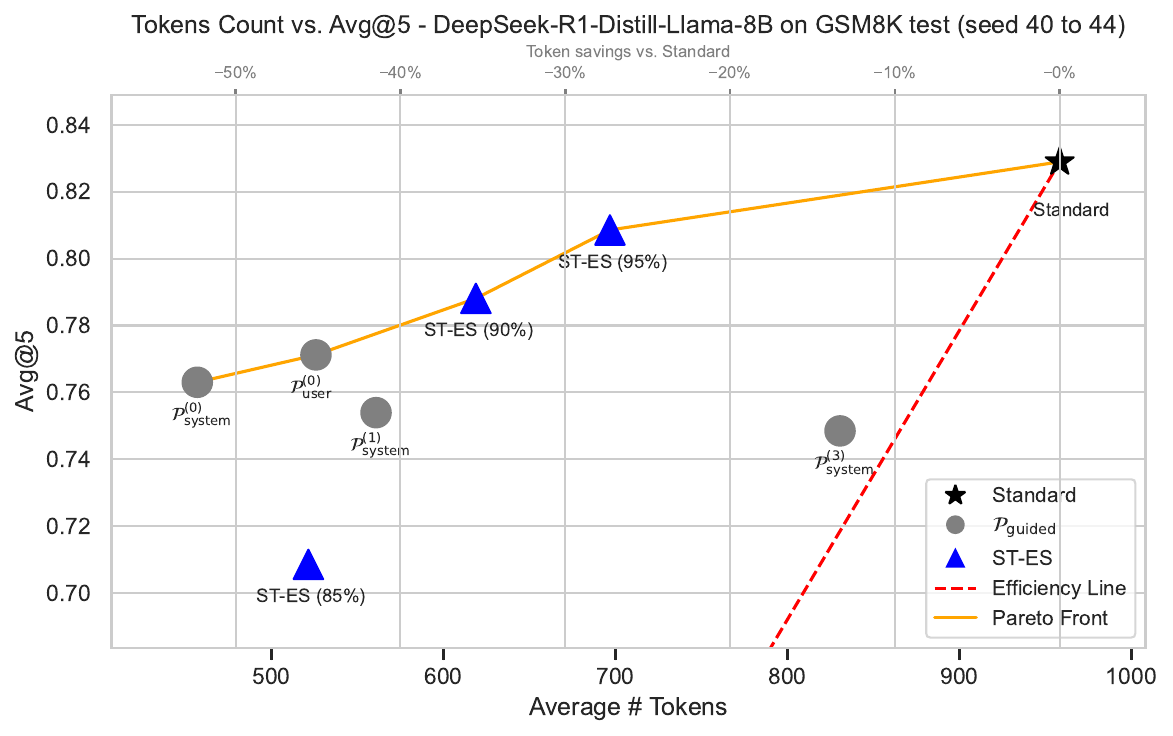}
        \caption{DS-Llama8B on GSM8K}
        \label{fig:DS8B-GSM8K-STES} 
    \end{subfigure}
    \hfill
    \begin{subfigure}{0.33\linewidth}
        \centering
        \includegraphics[width=\linewidth]{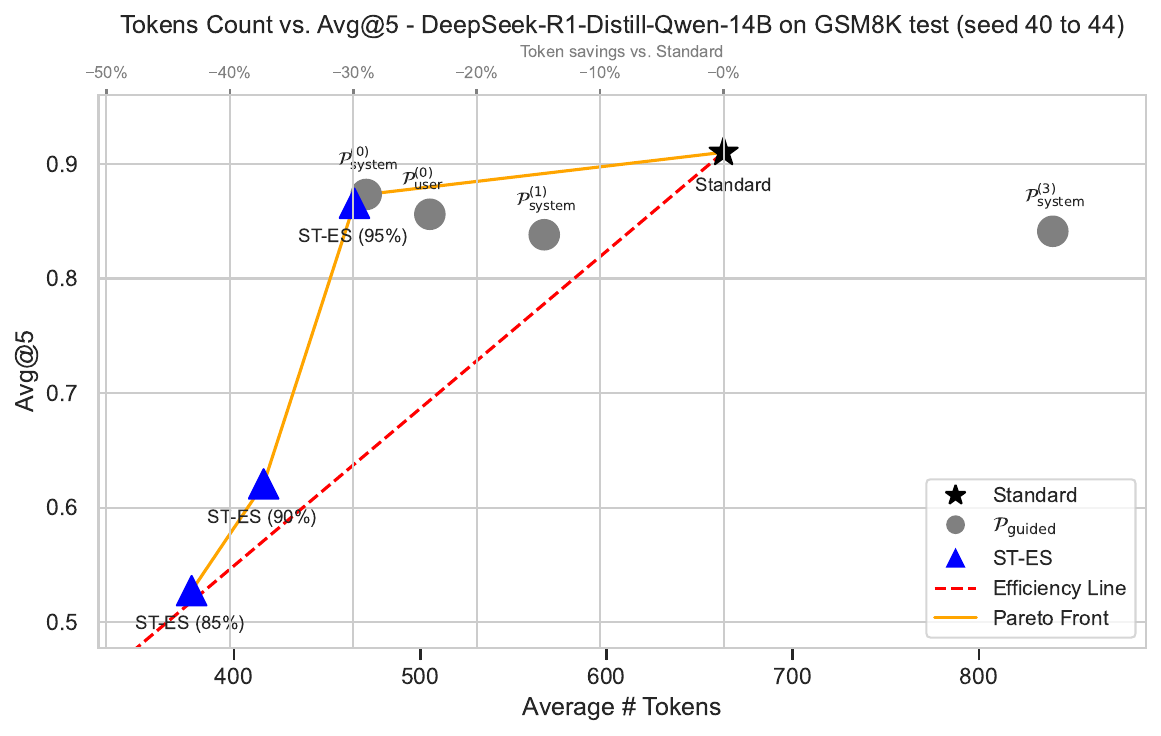}
        \caption{DS-Qwen14B on GSM8K}
        \label{fig:DS14B-GSM8K-STES}
    \end{subfigure}
    \hfill
    \begin{subfigure}{0.33\linewidth}
        \centering
        \includegraphics[width=0.99\linewidth]{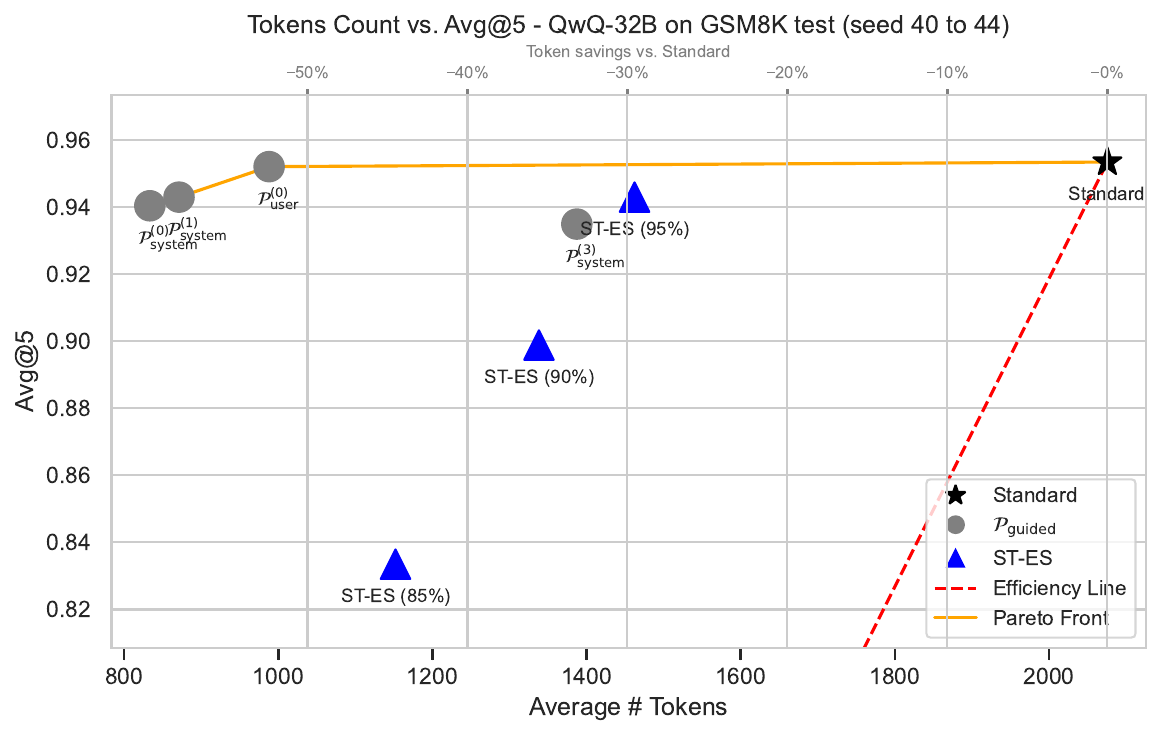}
        \caption{QwQ-32B on GSM8K}
        \label{fig:QwQ32B-GSM8K-STES}
    \end{subfigure}
    \begin{subfigure}{0.33\linewidth}
        \centering
        \includegraphics[width=\linewidth]{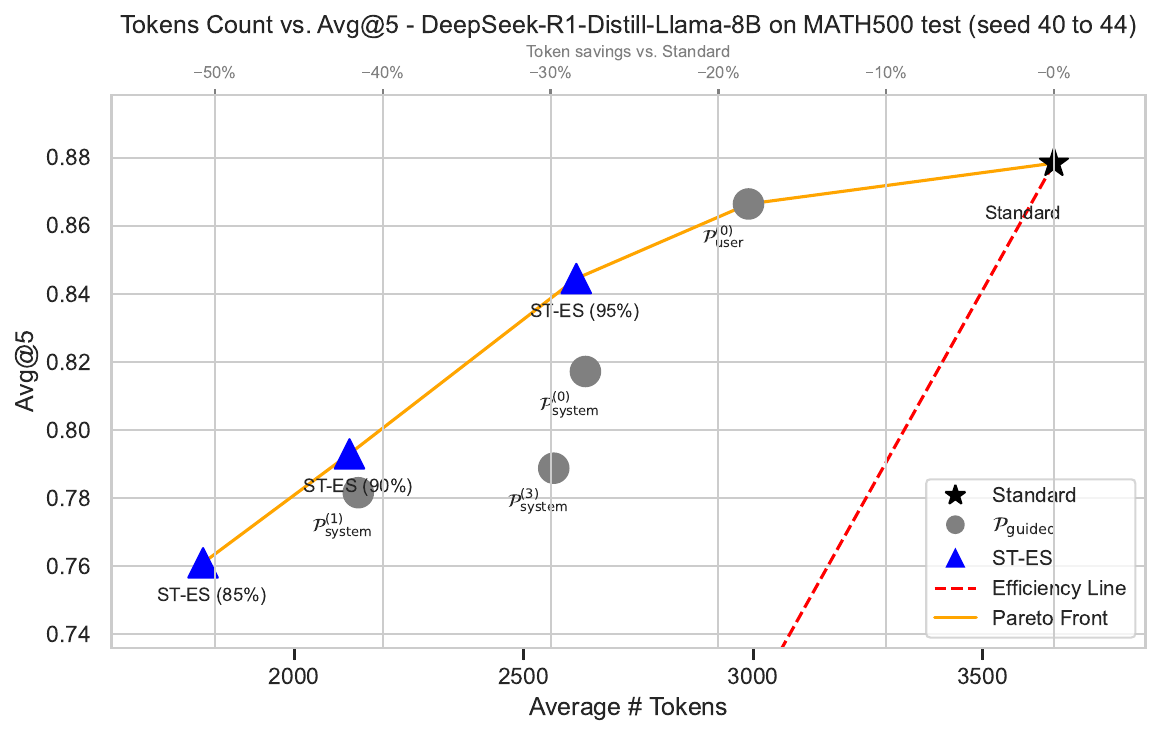}
        \caption{DS-Llama8B on MATH500}
        \label{fig:DS8B-MATH500-STES} 
    \end{subfigure}
    \hfill
    \begin{subfigure}{0.33\linewidth}
        \centering
        \includegraphics[width=\linewidth]{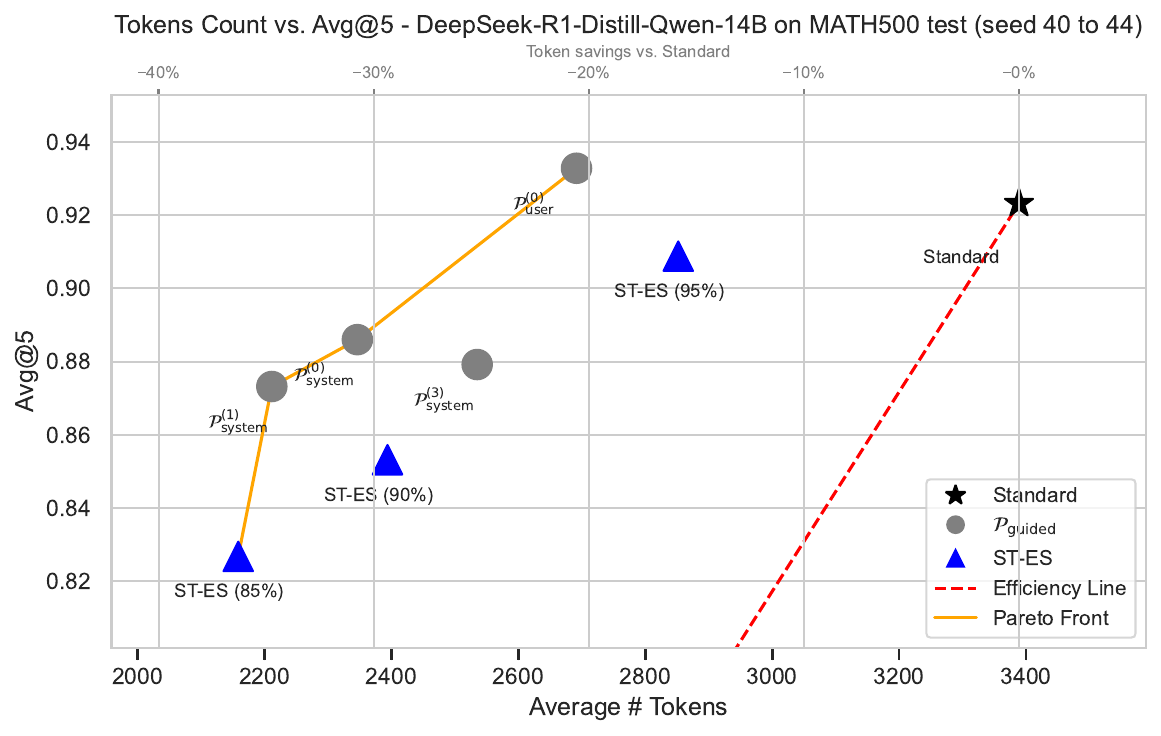}
        \caption{DS-Qwen14B on MATH500}
        \label{fig:DS14B-MATH500-STES}
    \end{subfigure}
    \hfill
    \begin{subfigure}{0.33\linewidth}
        \centering
        \includegraphics[width=0.99\linewidth]{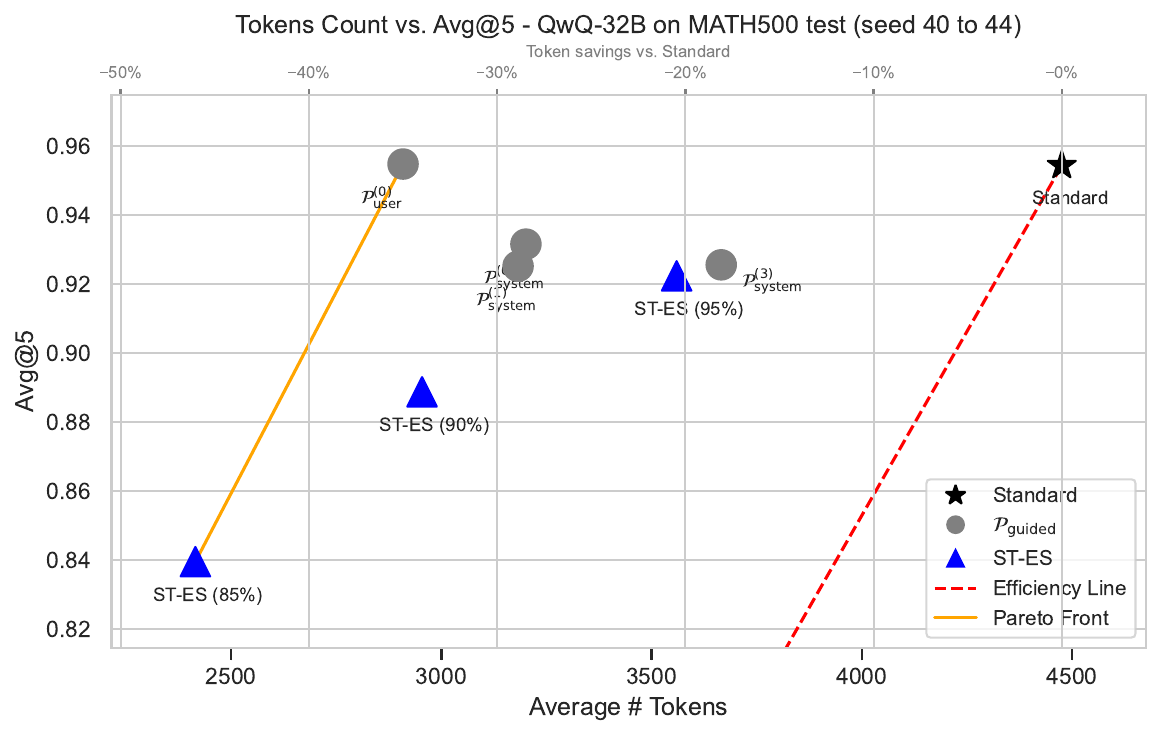}
        \caption{QwQ-32B on MATH500}
        \label{fig:QwQ32B-MATH500-STES}
    \end{subfigure}
    \\
    \begin{subfigure}{0.33\linewidth}
        \centering
        \includegraphics[width=\linewidth]{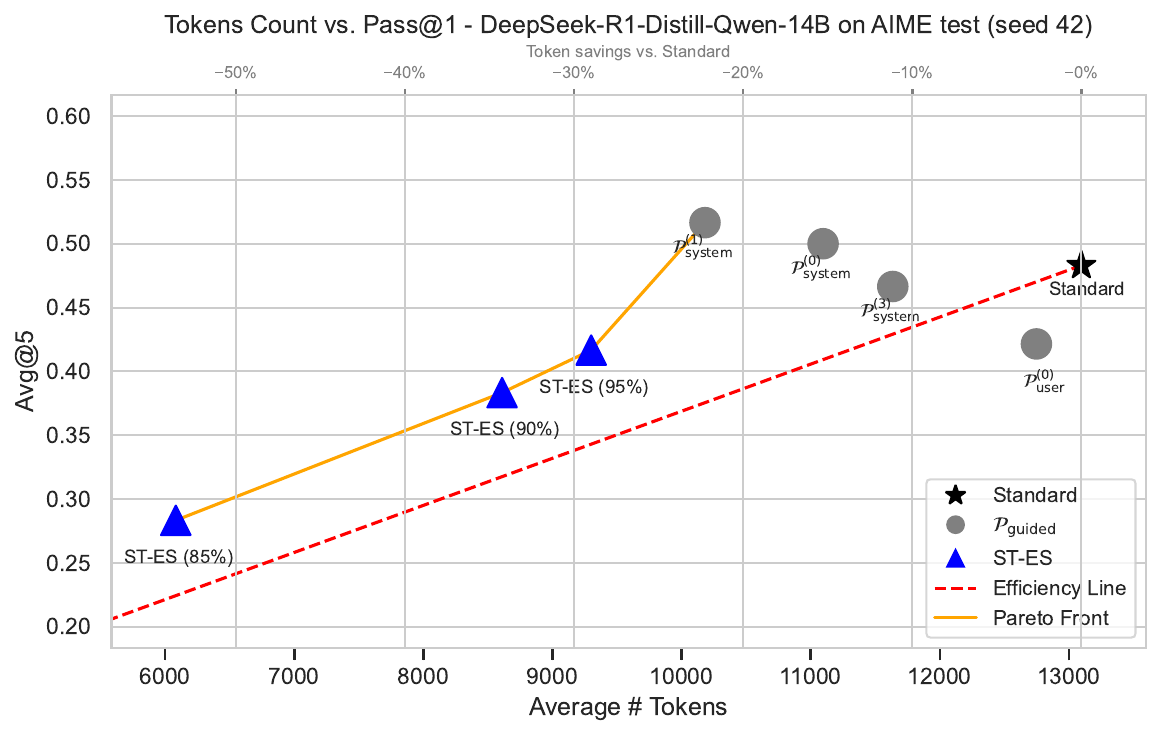}
        \caption{\small DS-Qwen14B on AIME}
        \label{fig:DS14B-AIME-STES} 
    \end{subfigure}
    \hfill
    \begin{subfigure}{0.33\linewidth}
        \centering
        \includegraphics[width=\linewidth]{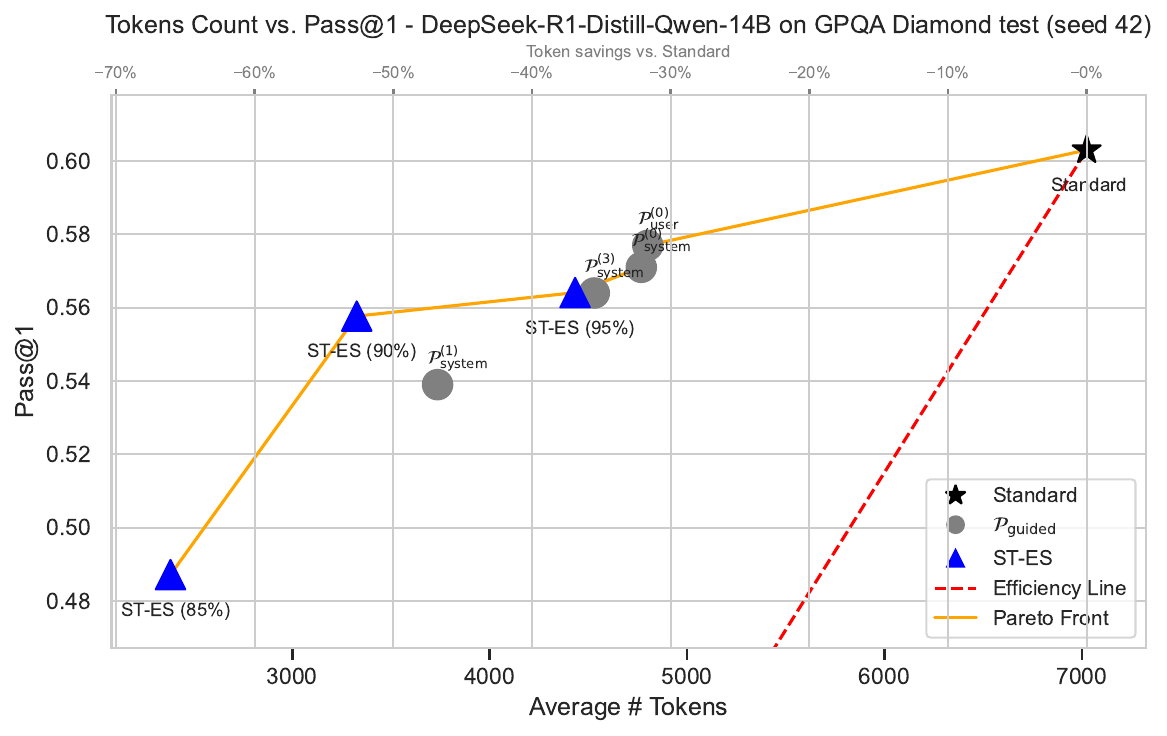}
        \caption{\small DS-Qwen14B on GPQA}
        \label{fig:DS14B-GPQA-STES}
    \end{subfigure}
    \hfill
    \begin{subfigure}{0.33\linewidth}
        \centering
        \includegraphics[width=\linewidth]{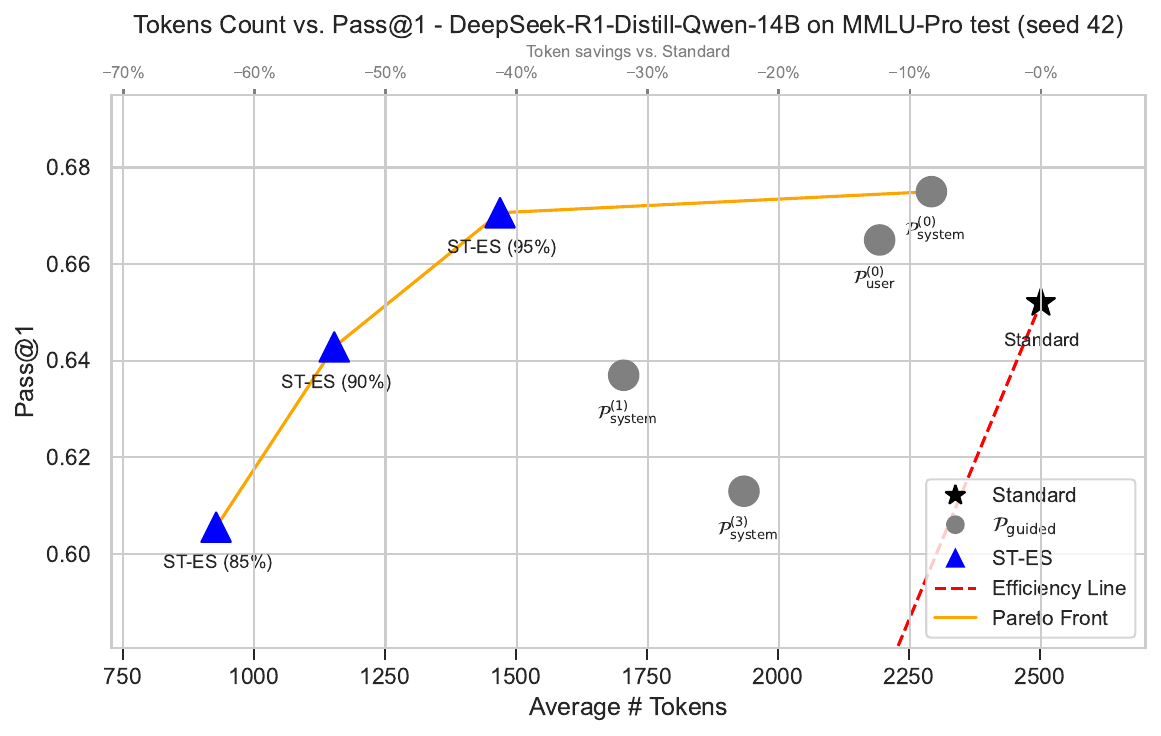}
        \caption{\small DS-Qwen14B on MMLU-Pro}
        \label{fig:DS14B-MMLU-STES}
    \end{subfigure}
    \vspace{-0.5cm}
    \caption{\small Number of Tokens vs. Avg@$5$ and Pass@$1$ - $\mathcal{P_{\text{guided}}}$ Baselines vs. ST-ES criteria on test datasets - The efficiency lines in red highlight the configurations that improve the efficiency relative to the standard inference, while the Pareto frontiers in yellow show the most efficient approaches. ST-ES achieved up to $20-50$\% token-count saving with minimal accuracy loss (from +2\% to -12\% accuracy).}
    \label{fig:st-es-performance}
    \vspace{-0.5cm}
\end{figure*}


\vspace{-0.1cm}

Next, we show in this section that Step-Tagging modules can effectively be used as an early-stopping criteria. Figure \ref{fig:st-es-performance} presents the average token count against the accuracy for the selected configurations. Each plot compares the performance trade-offs between the prompting baselines and the ST-ES criteria. Tables \ref{tab:full_results_5_seeds} and \ref{tab:other_dataset_st_es} in the Appendix \ref{sec:appendix-st-es-performance} report the quantitative metrics of the baselines and our approach on the three models, and five datasets.

\noindent \textbf{$\mathcal{P_{\text{guided}}}$ baselines.} We first notice that simple instruction on the models results in strong token-reduction, achieving $20$\% to $60$\% saved tokens across configurations. Specifically, the baselines achieve much better results on QwQ-32B, and the system-prompt variants generally lead to more token-reduction for the Deepseek models.

\noindent \textbf{Performance of ST-ES.}
Next, we observe that our ST-ES criteria achieved more efficient generation, with all ES-ST settings lying on the left side of the \emph{efficiency line} (in red) compared to the standard inference for all models. Further, ST-ES outperforms most $\mathcal{P_{\text{guided}}}$ baselines for both Deepseek models. Our ST-ES criteria is performing well on the DS-Llama8B model on both GSM8K and MATH500 since almost all ST-ES configurations lies on the Pareto front. For instance, on MATH500 (Figure \ref{fig:DS8B-MATH500-STES}), ST-ES 85\% achieved approximately the same token reduction as $\mathcal{P^{\text{(0)}}_{\text{system}}}$ and $\mathcal{P^{\text{(3)}}_{\text{system}}}$ (around 30\%), while achieving higher accuracy. 

\textbf{Generalization to other tasks.} To assess robustness beyond MATH500 and GSM8K dataset, Figures \ref{fig:DS14B-AIME-STES}-\ref{fig:DS14B-MMLU-STES} show the performance of ST-ES on AIME, GPQA-Diamond and MMLU-Pro on DS-Qwen14B. Our results indicate that ST-ES scales well on more complex tasks. All ST-ES lies on the Pareto front, offering 30 to 55\% token reduction with +2\% to -5\% accuracy for moderate configurations. Moreover, it shows that ST-ES remains effective across domains. 

\textbf{Robustness of ST-ES.} Appendix \ref{sec:cost-trade-off} shows that the training cost of ST-ES are fully recovered by the efficiency gains obtained. As well, Appendix \ref{sec:impact-bert-step-tagging} shows that our trained tagging modules approximates well the GPT-4o-mini annotation.


\vspace{-0.2cm}

\section{Conclusion}

\vspace{-0.15cm}

This work offers a novel view on both monitoring and efficiency of LRMs. We demonstrate that users can effectively track the reasoning flow of LRMs using our \emph{Step-Tagging} framework, paving the way for more work on the monitoring of reasoning steps. Furthermore, we show that tracking the step-type in the generation of LRMs can lead to reliable early-stopping criterion. Our framework can enhance the control of the generation of RLMs enabling a significant token-count saving (20-50\%) while preserving performance. 

\newpage

\section*{Acknowledgment}

This work has been partially supported by the 6G-XCEL project (grant agreement 101139194), funded by the EU Horizon Europe program. 

This research was partly supported by the ADAPT Research Centre. The ADAPT Centre for Digital Content Technology is funded under the Research Ireland’s Research Centres Programme (Grant 13/RC/2106 II) and is co-funded under the European Regional Development Funds.

\bibliography{camera_ready}
\bibliographystyle{icml2026}

\newpage
\appendix
\onecolumn

\section*{LLM Usage}

We acknowledge the use of Large Language Models for the purpose of our experimentation in our paper. Specifically, as stated in Section \ref{sec:step-tagging-module}, we relied on \texttt{GPT-4o-mini} to set our \emph{ReasonType} taxonomy. This approach is borrowed from work on behavior analysis of LLMs, such as \citet{galichin2025icoveredbaseshere, kuznetsov2025featurelevelinsightsartificialtext}.

\section*{Appendix}

\section{Future work}


While our ST-ES framework offers significant efficiency gains while maintaining the model's performance, a few limitations should be acknowledge. First, we introduced the \emph{ReasonType} taxonomy in Section \ref{sec:monitoring-step-tagging}, and demonstrate though experiments and ablations study that this taxonomy enable meaningful dynamic step-level monitoring of LRM generation. However, future work should focus on applying ST-ES on other taxonomies \citep{galichin2025icoveredbaseshere,marjanovic2026deepseekr1thoughtologyletsthink, minegishi2025topologyreasoningunderstandinglarge, venhoff2025basemodelsknowreason}.

Further, ST-ES requires calibration to find the most adapted inference configurations (step-types and value of $\delta$). We found across datasets and models that Verification and self-reflection steps with values of $\delta < 10$ are offering satisfying efficiency trade-offs. And we found that ST-ES is more practical than token-count baselines. Yet, the models are sensible to the value of $\delta$. Future work should focus on experimenting dynamic values of $\delta$ to make the criteria even more agnostic.

\newpage

\section{Construction and Validation of the ReasonType taxonomy} \label{app:constr-valid-taxonomy}

This appendix presents the methodology that we adopted to create the ReasonType taxonomy. Importantly, this section demonstrates the robustness of our annotation approach using our taxonomy, further validating our overall methodology.

\subsection{Construction of the taxonomy} \label{app:construction-taxonomy}

In Section \ref{sec:step-tagging-module}, we provided a high-level description of our methodology to generate a taxonomy of reasoning step. In this section, we will describe our methodology in more details. 

\textbf{Methodology.} Our goal in constructing the \emph{ReasonType} taxonomy is to observe the behavior of LRMs and look at aggregating similar types together to follow the generation of LRMs closely. Because no fine-grained taxonomy of reasoning steps exists, we used an open-ended labeling procedure: we prompted a model to generate free-form step-type labels, and manually merged common labels together. We took inspiration from previous work, who relied on the summarization and behavior detection capabilities of strong models such as \texttt{GPT-4o-mini}, as this method has proven to be effective \citep{galichin2025icoveredbaseshere, kuznetsov2025featurelevelinsightsartificialtext}. 

\begin{table}[h]
\tiny
\centering
\begin{tabular}{lccc}
\toprule
\textbf{Dataset} & \textbf{Model} & \textbf{\# Steps} & \textbf{\# Unique Tags} \\ 

\midrule

\multirow{2}{*}{MATH500} & \deepseekLlama{} & 3,840 & 162 \\
& \QwQLarge{} & 3,057 & 179 \\

\bottomrule
\end{tabular}
\caption{Step for taxonomy generation}
\label{tab:taxonomy-generation-traces}
\end{table} 

We first generated $100$ reasoning traces from the MATH500 train dataset, using \deepseekLlama{} and \QwQLarge{}, to obtain a pool of reasoning steps ($20$ samples from each complexity level). We obtained a pool of $6,897$ reasoning steps (see Table \ref{tab:taxonomy-generation-traces}). Each step is then passed to \texttt{GPT-4o-mini} using our Taxonomy prompt, presented in Figure \ref{fig:prompt_taxonomy} below. We obtained an open-ended label for each step-type. 

\begin{figure}[h]
\centering
\footnotesize
\begin{adjustbox}{max width=\textwidth}
\begin{tcolorbox}[colback=gray!5, colframe=black, title=Taxonomy Prompt, fonttitle=\bfseries]

Below is a reasoning trace of a reasoning language model, split by steps. In these examples, can you please identify the different type of steps? Suggest some reasoning-type labels for each of them. \\
- Step 1: \{step\_1\} \\
- \textbf{[...]} \\
- Step t: \{step\_t\}

\end{tcolorbox}
\end{adjustbox}
\caption{Prompt used to generate the Taxonomy}
\label{fig:prompt_taxonomy}
\end{figure}

\begin{figure}[h]
    \centering
    \includegraphics[width=1.0\linewidth]{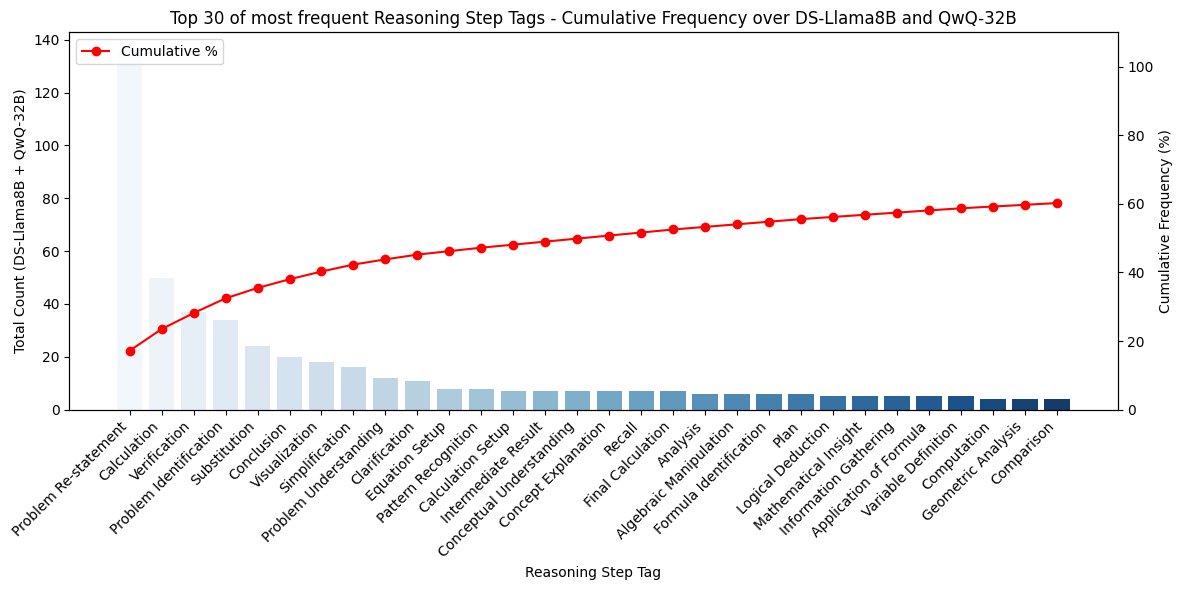}
    \caption{Most frequent labels obtained from open-end label generation}
    \label{fig:frequency-most-occurent-labels}
\end{figure}

Figure \ref{fig:frequency-most-occurent-labels} presents the most frequent labels generated. We observe that, even though our prompting was open-ended (without particular instruction for the model to generate certain type of labels), some labels were obtained frequently, such as \emph{Problem Identification/Re-Statement}, \emph{Substitution}, or \emph{Verification}. We manually inspected the labels obtained, and merged labels that have a common signification. Table \ref{tab:example-of-tag-merged} illustrates several examples. 

\textbf{Takeaways.} While the ReasonType taxonomy might not be optimal, we showed that it reflects the different steps of the model, and offers enough consistency and granularity to analyze the reasoning process of LRMs and to train accurate classifiers for our study. It also echoes previous work, and the categories overlaps with existing taxonomies \citep{galichin2025icoveredbaseshere,marjanovic2026deepseekr1thoughtologyletsthink, minegishi2025topologyreasoningunderstandinglarge, venhoff2025basemodelsknowreason}. Furthermore, the size of our taxonomy (13-14 step-types) is supported by the finding of \citet{venhoff2025basemodelsknowreason}: \emph{"we find [...] that reasoning mechanisms are reasonably well represented using 10 to 20 categories."}.

\textbf{Limitations.} To over-come the domain dependency, we could rely on the OpenThoughts-114k dataset. Indeed, it seems to provide a diverse set of possible behavior for the models. However, it would come at a high processing cost.

\begin{table}[h]
\small
\centering
\begin{tabular}{cp{6cm}p{6cm}}
\toprule
\textbf{Tags from ReasonType} & \deepseekLlama{} & \textbf{QwQ-32B} \\ 

\midrule

\multirow{2}{*}{Problem Re-statement / Setup} & Problem Re-statement, Problem Identification, Clarification & Problem Re-statement, Coordinate Setup, Step-by-Step Breakdown \\

\midrule

\multirow{1}{*}{Definition Recall} & Recall of Relevant Concepts, Definition Recall & Recall, Rule Recall, Definition Explanation \\

\midrule

\multirow{2}{*}{Formula Substitution} & Value Substitution, Application of Formula, Calculation & Equation Setup, Equation Manipulation, Application of Formula \\

\midrule

\multirow{2}{*}{Exploration} & Approach Exploration, Exploration of Alternatives & Exploration \\

\midrule

\multirow{2}{*}{Self-Talk} & Procedure Explanation, Plan of Action, Comparison of Options & Confirmation, Reflection, Logical Breakdown \\

\midrule

\multirow{2}{*}{Verification} & Confirmation, Verification & Backward Calculation, Example Verification, Assumption Checking \\

\midrule

\multirow{1}{*}{Final Answer} & Final Evaluation, Final Calculation & Final Evaluation, Final Calculation \\

\bottomrule
\end{tabular}
\caption{Example of categories from the ReasonType Taxonmy obtained by merging labels from the annotation - each set of labels were generated using our prompt and sampled reasoning steps, and we manually merged common labels to create the ReasonType taxonomy.}
\label{tab:example-of-tag-merged}
\end{table}

\clearpage

\newpage

\subsection{Labeling the reasoning traces} \label{sec:appendix-labeling-process}

Table \ref{tab:annotation-steps-data} presents statistics on the number of steps and \texttt{GPT-4o-mini} annotation for each models on the selected datasets. 

\begin{table}[h]
\tiny
\centering
\begin{tabular}{lccccccc}
\toprule
\textbf{Model} & \textbf{Dataset} & \textbf{\# Tok. / Steps} & \textbf{\# Steps / Sample} & \textbf{\# Steps} & \textbf{Runtime / Steps} & \textbf{Runtime / Sample} & \textbf{Total Runtime} \\ 

\midrule

\multirow{2}{*}{\texttt{DS-Llama-8B}} & GSM8K & 34.58 & 23.64 & 70,911 & 0.87 & 20.71 & 62,143 \\
& MATH500 & 34.11 & 96.27 & 96,270 & 0.88 & 84.86 & 84,861 \\

\midrule

\multirow{5}{*}{\texttt{DS-Qwen-14B}} & GSM8K & 36.64 & 14.01 & 42,017 & 0.86 & 12.09 & 36,293 \\
& MATH500 & 34.33 & 94.84 & 94,838 & 0.85 & 80.26 & 80,263 \\
& AIME 22-24 & 27.74 & 469.29 & 42,236 & 0.85 & 398.68 & 35,881 \\
& GPQA-Diamond & 38.96 & 186.51 & 36,928 & 0.84 & 156.05 & 30,898 \\
& MMLU-Pro & 35.86 & 64.59 & 90,428 & 0.87 & 56.42 & 78,985 \\

\midrule

\multirow{2}{*}{\texttt{QwQ-32B}} & GSM8K & 45.59 & 40.43 & 121,288 & 0.87 & 35.33 & 105,977 \\
& MATH500 & 38.97 & 117.28 & 117,283 & 0.83 & 97.49 & 97,494 \\

\bottomrule
\end{tabular}
\caption{Avg. \# of steps and annotation runtime per sample - MATH500 and GSM8K are the trained dataset ($1,000$ and $3,000$ samples, respectively). AIME, GPQA and MMLU are the full dataset ($90$, $198$ and $1,400$ samples, respectively). Runtime in seconds.}
\label{tab:annotation-steps-data}
\end{table}

\vspace{-0.5cm}

\subsection{Reliability of \texttt{GPT-4o-mini} as an annotator} \label{sec:appendix-reliability-annotation}

\textbf{Claim.} Our framework rely on the annotation capability of the \texttt{GPT-4o-mini} model. Since the model is large and achieved great performance on a range of tasks, we assumed that the model is able to provide us with labels of good quality. In this section, we will observe and analyse the reliability of the annotation of the steps by the \texttt{GPT-4o-mini} model.

\textbf{Methodology.} To verify our claim, we sampled $1,000$ reasoning steps of DS-Qwen14B model from its inference on the MATH500 dataset. We then annotated each steps $5$ times using \texttt{GPT-4o-mini}, and observe the agreement of each annotation. In addition, we also compared the annotation agreement between the \texttt{GPT-4o-mini} model, and 3 additional models: \texttt{GPT-4o} (a larger, closed-source model), \texttt{llama-3-3-70b-instruct}, and \texttt{Mixtral-8x22B-Instruct-v0.1} (both open-source and smaller relatively to the two model selected). This setup allows us to assess both the internal consistency of \texttt{GPT-4o-mini} and its alignment with other model annotators.

\textbf{Experimental Design.} We refer to self-model agreement as the \emph{Inner-model agreement} (agreement between different runs of the same model on the same reasoning steps). The inner-model agreement is measured using the Fleiss' kappa metric. Indeed, the Fleiss' kappa measure the agreement between more than $2$ annotators, making it suitable to compare many annotation trials \citep{Moons_2025}. Furthermore, we call agreement between two different model the \emph{Inter-model agreement}. To compute this, we first selected the most consistent label generated across the different runs for each model, and we measured the Cohen's Kappa \citep{badshah2025referenceguidedverdictllmsasjudgesautomatic}. For this experiment, we used the OpenAI default decoding parameters for the GPT models. Same parameter was set for the other models selected.

\begin{table}[h]
\footnotesize
\centering
\begin{tabular}{lcccc}
\toprule
\textbf{Models} & GPT-4o-mini & GPT-4o & Llama-3-3-70b & Mixtral-8x22B \\ 

\midrule

GPT-4o-mini & \textbf{0.780} & 0.601 & 0.457 & 0.392 \\
GPT-4o &  & \textbf{0.799} & 0.445 & 0.384 \\
Llama-3-3-70b &  &  & \textbf{0.722} & 0.398 \\
Mixtral-8x22B &  &  &  & \textbf{0.587} \\

\bottomrule
\end{tabular}
\caption{Agregation metrics of the annotation process - $1,000$ samples reasoning steps from the \deepseekQwen{} model on the MATH500 dataset -  \textbf{Fleiss' kappa} in bold (diagonal - inner-model aggregation) - \textbf{Cohen's kappa} otherwise (inter-model aggregation)}
\label{tab:reliability-annotation}
\vspace{-0.5cm}
\end{table}

\textbf{Internal consistency of \texttt{GPT-4o-mini}.} The Fleiss' Kappa score of $0.780$ (see Table \ref{tab:reliability-annotation} in the diagonal) indicates high agreement among multiple independent runs of \texttt{GPT-4o-mini} on the same reasoning steps. This demonstrates that the model produces stable and consistent annotations when prompted repeatedly. We observe that the Fleiss' Kappa score seems to decrease as the model size becomes smaller.

\textbf{\texttt{GPT-4o-mini} against other models.} When comparing \texttt{GPT-4o-mini} with other models, we observe significant agreement with \texttt{GPT-4o}, with a Cohen's Kappa of $0.601$, and moderate agreement with smaller models ($0.457$ with Llama70b and $0.392$ with Mixtral). The higher agreement with \texttt{GPT-4o} suggests that larger models tend to produce more stable and higher-quality annotation, while smaller models exhibits more variability.

\textbf{Takeaway.} Overall, these results supports that \texttt{GPT-4o-mini} is a reliable annotator for reasoning steps. Indeed, for our annotation task, the model appears to be consistent accross multiple runs, and shows meaningful agreement with other strong models. The results obtained on smaller models confirms our decision of selecting \texttt{GPT-4o-mini} as a annotator to label the reasoning steps of LRMs.

\newpage

\subsection{Reason-Type, a robust taxonomy for identifying reasoning behaviors} \label{sec:appendix-validation-taxonomy-shuffled}


\textbf{Objective.} This ablation study is looking at further validating our ReasonType taxonomy. In other words, we are investigating whether our proposed taxonomy captures meaningful distinctions in reasoning steps. We are looking to demonstrate that: 

\begin{enumerate}[leftmargin=*, itemsep=0pt]

    \item The ReasonType taxonomy enable semantic distinction of the type of reasoning.

    \item Our annotation method with the \texttt{GPT-4o-mini} model, coupled with the ReasonType taxonomy, is a robust method to access to the ground-truth labels of the reasoning steps. 
    
\end{enumerate}

\textbf{Methodology.} To address our objective, we compare the performance of BERT classifiers across Original labels (OG - from \texttt{GPT-4o-mini} annotation using the ReasonType taxonomy), and shuffled labels for three step-types, namely: \emph{Verification}, \emph{Exploration} and \emph{Self-Talk}. For the shuffled labels version, we took the exact same proportion of positive labels as in Original datasets, and used random shuffle with a seed of $42$. Each experiment is run on the same training and held-out validation dataset, using the MATH500 training dataset on the DS-Qwen14B model. We trained binary step-taggers following the same training configuration (see Section \ref{sec:monitoring-step-tagging}). We set the label of selected class to $1$, and put other labels to $0$. To compare performances, we report both training loss, and classification metrics (precision and recall on both classes, along with macro and micro average.)

\begin{minipage}{0.4\linewidth}
    \textbf{Evaluation.} Figure \ref{fig:shuffled_training_loss} shows the training loss of the Original and Shuffled versions, for the three labels. Models trained on the Original labels presents significant lower losses, and are smoothly decreasing. It demonstrate that the Original datasets contains meaningful patterns between reasoning steps and their labels. In comparison, the models trained on shuffled labels present almost constant loss, relatively higher than the one from the Original labels.
\end{minipage}
\hfill
\begin{minipage}{0.6\linewidth}
    \centering
    \includegraphics[width=0.85\linewidth]{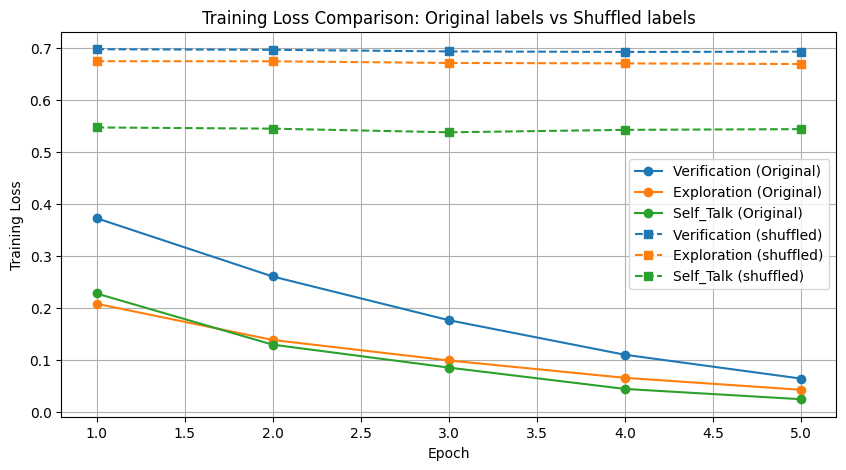}
    \captionof{figure}{\small Training losses - ReasonType vs. Shuffled labels}
    \label{fig:shuffled_training_loss}
\end{minipage}

\begin{figure}[h]
    \centering
    \begin{subfigure}{0.48\textwidth}
        \includegraphics[width=\linewidth]{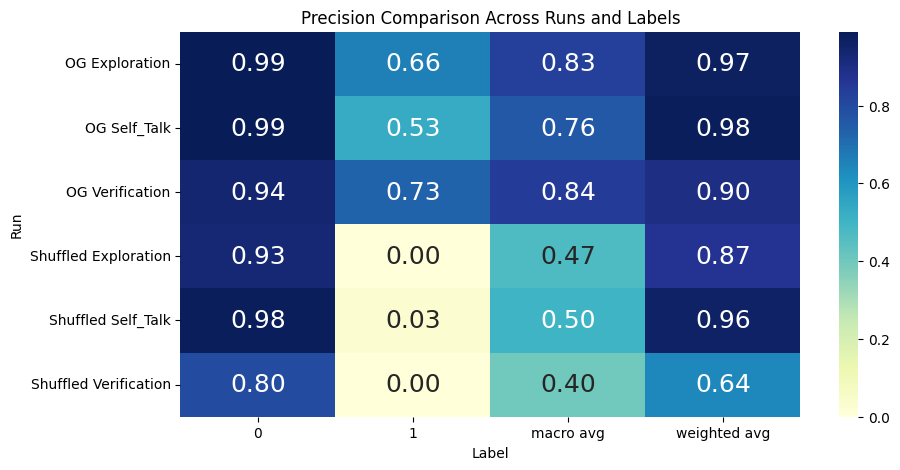}
        \caption{Precision}
        \label{fig:shuffled_precision}
    \end{subfigure}
    \hfill
    \begin{subfigure}{0.48\textwidth}
        \includegraphics[width=\linewidth]{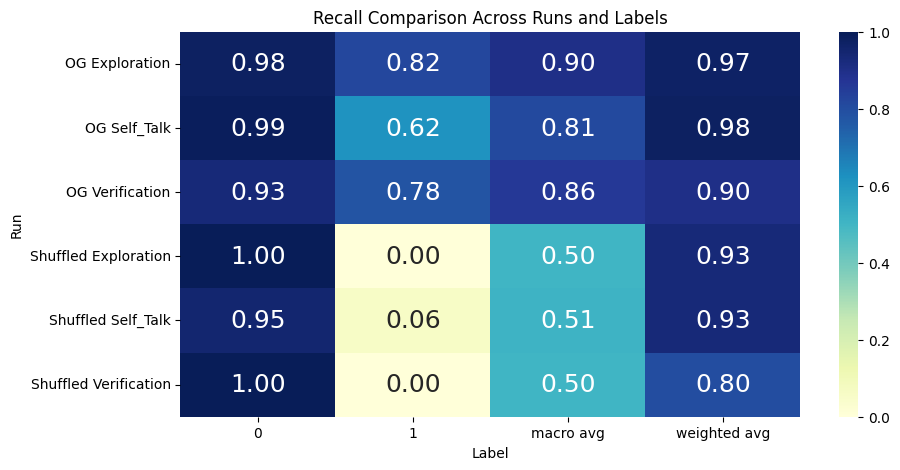}
        \caption{Recall}
        \label{fig:shuffled_recall}
    \end{subfigure}
    \vspace{-0.2cm}
    \caption{Precision and Recall - ReasonType vs. Shuffled labels}
    \label{fig:shuffled_metrics}
    \vspace{-0.2cm}
\end{figure}

Furthermore, Figures \ref{fig:shuffled_precision} and \ref{fig:shuffled_recall} show the Precision and Recall classification metrics on the testing dataset, respectively. For Original runs, both classes ($0$ and $1$) achieve good performance despite dataset imbalancity, with Macro average Precision and Recall lying between $0.76$ and $0.90$ across labels. In comparison, shuffled runs presents poor results, with models failing in predicting positive classes - Precision and Recall of class $1$ between $0.00$ and $0.06$. Along with the training loss, theses metrics highlight that the models trained on shuffles labels cannot learn meaningful relations between steps and labels. In comparison, Original labels (from the ReasonType taxonomy) resulted in satisfying model performance, and smooth training.

\textbf{Takeaway.} Overall, these results comforts our finding that the ReasonType taxonomy labels enable annotation methods to results in reasoning steps carrying semantic meaning.

\newpage

\subsection{Generalizability of the ReasonType taxonomy} \label{sec:appendix-generalization-taxonomy}

We proved in Appendix \ref{sec:appendix-reliability-annotation} that \texttt{GPT-4o-mini} is a reliable annotator model for our taxonomy and the task of tagging the steps. However, the taxonomy is generated from specific samples of specific models. We already proved to some extend that the Taxonomy is generalizable. Indeed, we trained our step-tagging classifier on reasoning steps obtained using DS-Qwen14B, a model that was not used to derive the taxonomy (see Appendix \ref{app:construction-taxonomy}). Moreover, we obtained satisfying performance of our step-tagging classifier on models and datasets that were not used to for training, namely DS-Llama8B and QwQ-32B on MATH500 and GSM8K, and DS-Qwen14B on GSM8K, AIME and MMLU-Pro (see Section \ref{sec:monitoring-step-tagging}).

\textbf{Claim.} To further validate our taxonomy, we will test its applicability to 2 additional reasoning models. Our claim is that if the annotation process from \texttt{GPT-4o-mini} results in accurate training of step-tagging classifiers, it means that the ReasonType taxonomy is applicable to other models, therefore generalizable. To evaluate this, we trained binary BERT classifiers -- similarly as Appendix \ref{sec:appendix-validation-taxonomy-shuffled} -- on four step-types, namely: \emph{Problem Re-statement}, \emph{Exploration}, \emph{Self-Talk} and \emph{Verification}.

\textbf{Methodology.} To verify our claim, we inferred 2 additional models on both MATH500 and GSM8K, specifically \texttt{microsoft/Phi-4-reasoning} and \texttt{Qwen/Qwen3-30B-A3B-Thinking-2507} since they are both reasoning models, and comes from additional providers or training methods. We then trained \textbf{$4$} BERT classifiers for each model and dataset. We used $500$ and $3,000$ samples from the MATH500 and GSM8K train datasets, respectively. 

\begin{figure}[h]
    \centering
    \begin{subfigure}{0.8\textwidth}
        \includegraphics[width=\linewidth]{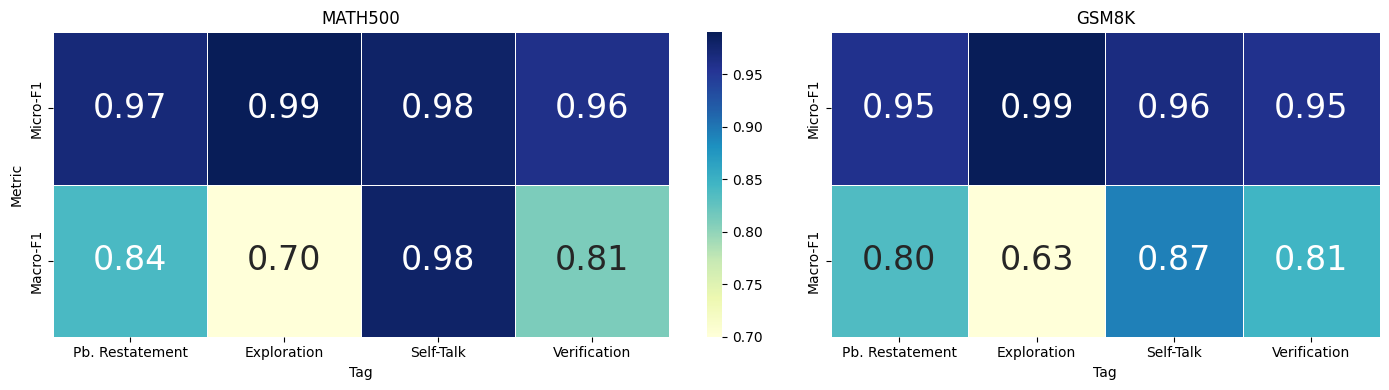}
        \caption{Phi-4}
        \label{fig:phi_4_reasoning}
    \end{subfigure}
    \hfill
    \begin{subfigure}{0.8\textwidth}
        \includegraphics[width=\linewidth]{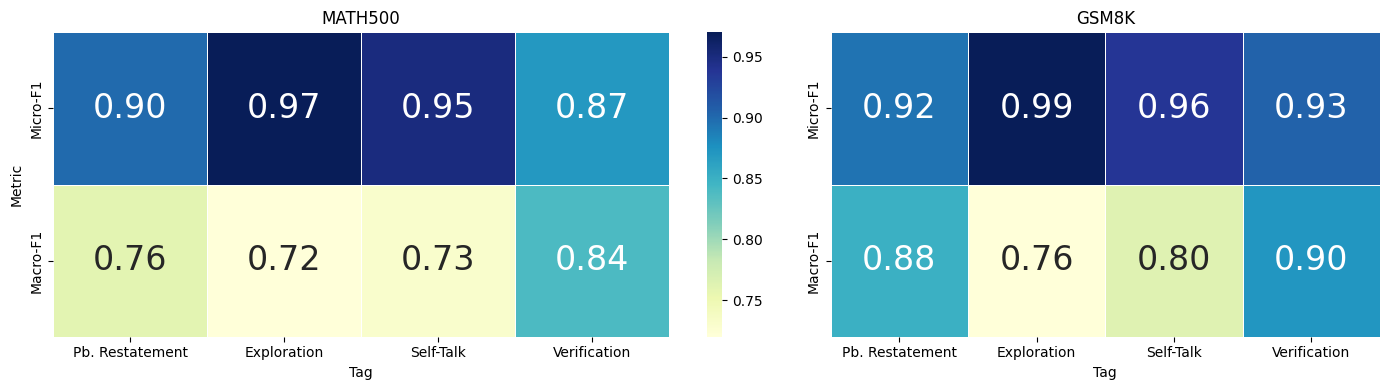}
        \caption{Qwen3-30B-A3B}
        \label{fig:Qwen3-30B-A3B}
    \end{subfigure}
    \vspace{-0.2cm}
    \caption{Performance of Step-Taggers - seed 42}
    \label{fig:performance_new_models}
    \vspace{-0.2cm}
\end{figure}

\textbf{Performances of the Step-Taggers.} We split the resulting annotated datasets following random 80:20 train/test split. Figure \ref{fig:phi_4_reasoning} and \ref{fig:Qwen3-30B-A3B} show the micro-F1 and macro-F1 for the Phi-4 and Qwen3-30B-A3B models on MATH500 and GSM8K, respectively. We observe satifying performances of Step-Taggers on both models. Specifically, on the \texttt{Phi-4} model, we obtained between $0.7$ to $0.98$ and $0.63$ to $0.87$ macro-F1 on MATH500 and GSM8K datasets, respectively. The lower macro-F1 on Exploration can be explained by its low representation in the datasets (around 1\% of labels in both datasets). 

Similarly, \texttt{Qwen3-30B-A3B} obtained satifying perfromance on both MATH500 and GSM8K, with $0.72$ to $0.84$ and $0.76$ to $0.90$ macro-F1, respectively. Performances of both models are comparable to results obtained on the DS-Qwen14B using the original dataset (see Appendix \ref{sec:appendix-validation-taxonomy-shuffled}).

\textbf{Takeaways.} We interpret the satisfying performances of the Step-Taggers trained on other models as further validating the applicability of our taxonomy to other models. Indeed, these models were not used to create our \emph{ReasonType} taxonomy, but still resulted in step-type identification performance of BERT classifiers trained on their reasoning traces using our methodology.

\newpage

\section{Step-Tagging Early-Stopping algorithm} \label{sec:appendix-es_algorithm}

Algorithm \ref{alg:st_es_algorithm} lists the Step-Tagging Early-Stopping criteria. The user needs to define a constraint \{$\tau*$, $\delta$\}, and input a Binary Step-Tagger $\phi_{\tau^*}$, which returns 1 if the step tag is $\tau^*$ and 0 otherwise. If the constraint breaks, the algorithm stops the generation, and prompts the model with $\mathcal{P}_{\text{exit}}$ to give the current best answer.

\begin{algorithm}[h]
\footnotesize
\caption{Step-Tagger Early-Stopping}
\label{alg:st_es_algorithm}
\begin{algorithmic}[1]
\REQUIRE Prompt $x$; reasoning delimiter $\alpha \in V$; max steps $T_{\max}$; Reasoning Language Model $\mathcal{M}$; tokenizer $\mathcal{T}$; EOS token $\gamma$; Constraint $\{\tau^*, \delta\}$; Binary Step-Tagger $\phi_{\tau^*}$; Early-Exit Prompt $\mathcal{P}_{\text{exit}}$
\STATE $y \gets \mathcal{T}(x)$ \COMMENT{Tokenize the input}
\STATE $S_{\text{running}} \gets [\,]$ \COMMENT{Initialize output}
\STATE $t \gets 0$
\STATE $f_{\tau^*} \gets 0$ \COMMENT{Initialize frequency count}
\WHILE{$c_{\tau^*}(S_{\text{running}}, \delta)$}
    \STATE Generate step $s_i$ using $\mathcal{M}$, $\alpha$ \COMMENT{Generate until constraint breaks}
    \STATE $y \gets s_i$
    \IF{$\phi_{\tau^*}(s_i)$}
        \STATE $f_{\tau^*} \gets f_{\tau^*} + 1$ \COMMENT{Increase the counter}
    \ELSE
        \STATE Continue the generation
    \ENDIF
    \STATE $t \gets t + 1$
\ENDWHILE
\STATE $y \gets \mathcal{M}(y + \mathcal{P}_{\text{exit}})$ \COMMENT{Infer $\mathcal{M}$ with the early-exit prompt}
\RETURN $y$
\end{algorithmic}
\end{algorithm}

\begin{figure}[h]
    \centering
    \includegraphics[width=1.0\linewidth]{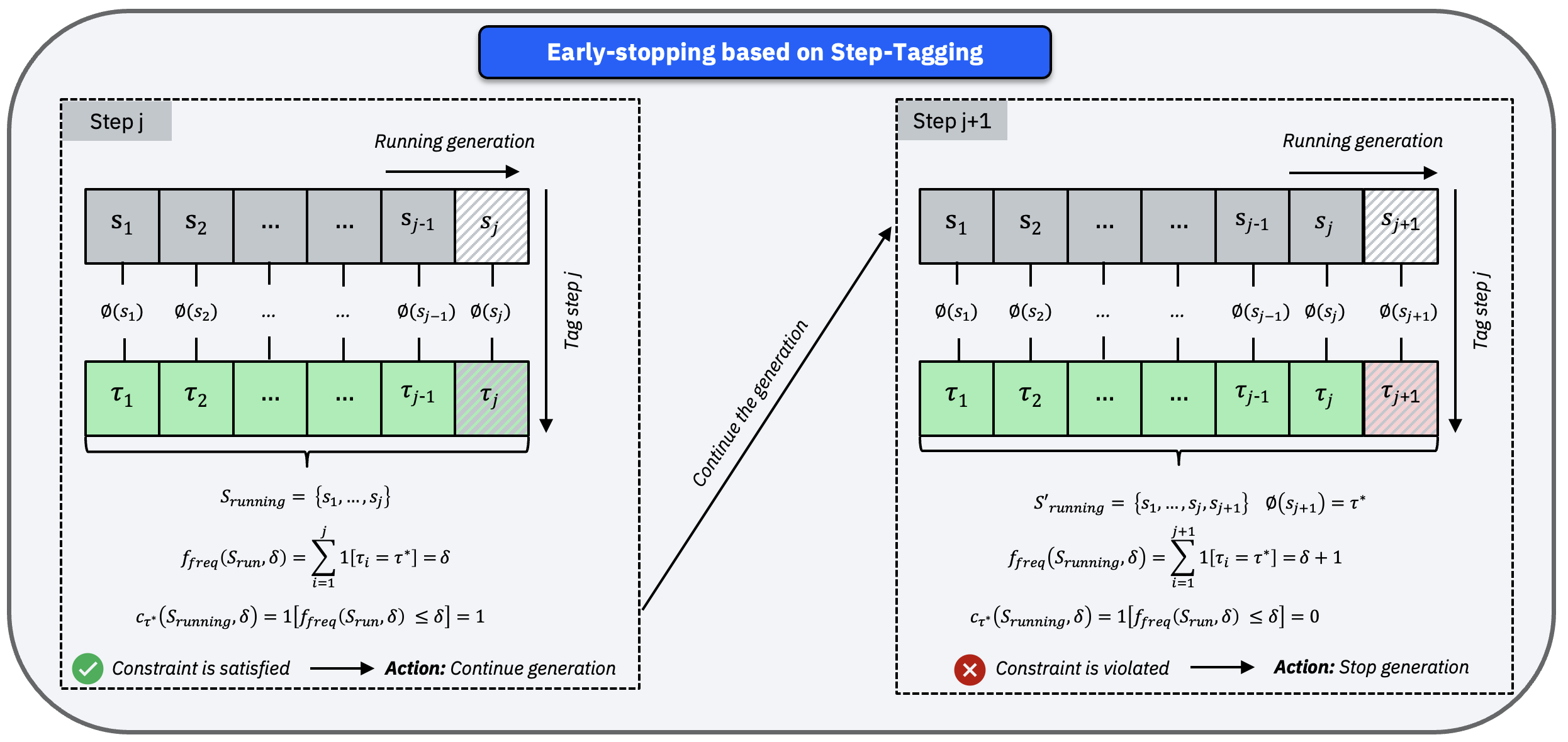}
    \caption{Illustration of early-stopping based on Step-Tagging}
    \label{fig:illustration-es-step-tagging}
\end{figure}

\newpage

\section{Experimental Design} \label{sec:appendix-experimental-design}

\textbf{Datasets.} Table \ref{tab:reasoning-datasets} summarizes the datasets and metrics we used to evaluate our selected reasoning models.

\begin{table}[h]
    \small
    \centering
    \begin{tabular}{ccccccc}
    \toprule
    \multirow{2}{*}{\textbf{Domain}} & \multirow{2}{*}{\textbf{Ref.}} & \multirow{2}{*}{\textbf{Name}} & \multirow{2}{*}{\textbf{Train Size}} & \multirow{2}{*}{\textbf{Test Size}} & \multicolumn{2}{c}{\textbf{Type of evaluation}} \\
    \cmidrule(lr){6-7} 
     &  &  &  &  & Math-Verify & MCQ Prompting \\
    \midrule
       \multirow{3}{*}{Maths} & \cite{cobbe2021trainingverifierssolvemath} & GSM8K & $3,000$ & $1,329$ & \cmark &  \\
        & \cite{hendrycks2021measuringmathematicalproblemsolving} & MATH500 & $1,000$ & $500$ & \cmark &  \\
        & \cite{aime2025dataset} & AIME 22-24 & $30$ & $60$ & \cmark &  \\
    \midrule
        \multirow{2}{*}{In-domain} & \cite{rein2023gpqagraduatelevelgoogleproofqa} & GPQA-Diamond & $40$ & $158$ &  & \cmark \\
        & \cite{wang2024mmluprorobustchallengingmultitask} & MMLU-Pro & $240$ & $1,200$ &  & \cmark \\
    \bottomrule
    \end{tabular}
    \caption{Overview of selected reasoning datasets and metrics}
    \label{tab:reasoning-datasets}
\end{table}

\noindent \textbf{Evaluating LRMs.} To assess the model's performance on challenging reasoning tasks, the Avg@$k$, Pass@$k$, and Cons@$k$ are common metrics \citep{chen2021evaluatinglargelanguagemodels, chen2025reasoningerasurveylong, yu2025dapoopensourcellmreinforcement}. The Pass@$k$ measures the proportion of the samples where at least one of $k$ attempts leads to the correct answer, while the Cons@$k$ consider a sample correct if all $k$ attempts are correct. Since we are interested about both performance and robustness of our approach, we selected the Avg@$5$, the Pass@$5$ and the Cons@$5$ as the quantitative metrics. Assessing the performance of LRMs on mathematical questions is challenging. This is due to the open nature of the question. For our experiments, we selected the \emph{Math-Verify}\footnote{\url{https://github.com/huggingface/Math-Verify}} library which is a common metric to assess mathematical problems. It uses text extraction and formal verification. This metric also reported strong correctness compared to other evaluation methods such as Harness \citep{eval-harness} or Qwen-Math Verifier \citep{huang2025accuracyrobustnessstudyrule}.

\noindent \textbf{Inference setting.} To monitor the steps and intervene in the generation process, we assume that each model generates one token at a time, and we split the steps dynamically. However, for the purposes of our experiments instead of re-designing the generation process, we performed standard inference and applied our \emph{Step-Tagging} and \emph{Early-Stopping} algorithms \emph{offline}.

\textbf{Prompt-guided budget compression.} Figure \ref{fig:prompt_baselines} presents the Prompt-guided baselines that we used to evaluate our ST-ES framework. 

\begin{figure}[h]
\centering
\tiny
\begin{adjustbox}{max width=\textwidth}
\begin{tcolorbox}[colback=gray!5, colframe=black, title=Prompt Baselines $\mathcal{P}_{\text{guided}}$, fonttitle=\bfseries, left=1pt, right=1pt]

\begin{tcolorbox}[
  colback=white, 
  colframe=black!30, 
  title=User Prompt - $\mathcal{P}^{(0)}_{\text{user}}$, 
  coltitle=black!30!black,
  top=2pt,
  bottom=2pt,
  boxsep=2pt,
  width=\textwidth,
  left=2pt, 
  right=2pt
]
\textbf{User Prompt:} Please do not reason extensively, be succinct, and put your final answer within boxed\{\}. \{question\}
\end{tcolorbox}


\begin{tcolorbox}[colback=white, 
    colback=white, 
    colframe=black!60, 
    title=System Prompt $\mathcal{P}^{(0)}_{\text{system}}$,
    top=2pt,
  bottom=2pt,
  boxsep=2pt,
  width=\textwidth,
  left=2pt, 
  right=2pt
]
\textbf{System Prompt:} Respond concisely and confidently. Skip validations and over-verification steps.

\textbf{User Prompt:} \{question\}

\end{tcolorbox}

\begin{tcolorbox}[colback=white, 
    colback=white, 
    colframe=black!60, 
    title=System Prompt - $\mathcal{P}^{(1)}_{\text{system}}$,
    top=2pt,
  bottom=2pt,
  boxsep=2pt,
  width=\textwidth,
  left=2pt, 
  right=2pt
]
\textbf{System Prompt:} Respond concisely and confidently. Skip validations and over-verification steps. Here is an examples: Example 1: \{FS\_1\}

\textbf{User Prompt:} \{question\}

\end{tcolorbox}

\begin{tcolorbox}[colback=white, 
    colback=white, 
    colframe=black!60, 
    title=System Prompt - $\mathcal{P}^{(3)}_{\text{system}}$,
    top=2pt,
  bottom=2pt,
  boxsep=2pt,
  width=\textwidth,
  left=2pt, 
  right=2pt
]
\textbf{System Prompt:} Respond concisely and confidently. Skip validations and over-verification steps. Here are some examples:  Example 1: \{FS\_1\}  Example 2: \{FS\_2\}  Example 3: \{FS\_3\}

\textbf{User Prompt:} \{question\}

\end{tcolorbox}

\begin{tcolorbox}[colback=white, 
    colback=white, 
    colframe=black!80, 
    title=Example 1 - Verification step,
    top=2pt,
  bottom=2pt,
  boxsep=2pt,
  width=\textwidth,
  left=2pt, 
  right=2pt
]

Wait, let me double-check. If I plug in $x = -3$ into the denominator, $(-3)^2 + (-3) -6 = 9 -3 -6 = 0$. Yep, that works. For $x =2: 2^2 +2 -6 =4 +2 -6=0$. Correct. So both roots are valid. 

\end{tcolorbox}

\begin{tcolorbox}[colback=white, 
    colback=white, 
    colframe=black!80, 
    title=Example 2 - Verification step,
    top=2pt,
  bottom=2pt,
  boxsep=2pt,
  width=\textwidth,
  left=2pt, 
  right=2pt
]

Therefore, the graph of $y=2\/(x^2 +x -6)$ has vertical asymptotes at $x= -3$ and $x=2$, so that's two vertical asymptotes. I don't think there's any chance that I made a mistake here, but maybe I should check by graphing the function or plugging in values close to $-3$ and $2$ to see if the function does go to infinity.

\end{tcolorbox}

\begin{tcolorbox}[colback=white, 
    colback=white, 
    colframe=black!80, 
    title=Example 3 - Verification step,
    top=2pt,
  bottom=2pt,
  boxsep=2pt,
  width=\textwidth,
  left=2pt, 
  right=2pt
]

Another test with $n=3$. Let's compute manually. All non-empty subsets: Single elements: \{1\}, \{2\}, \{3\} with sums 1,2,3. Pairs: \{1,2\} → $2-1=1$\; \{1,3\} → $3-1=2$\; \{2,3\} → $3-2=1$. Triple: \{1,2,3\} → $3 -2 +1=2$. Total sum: $1+2+3 +1+2+1 +2 = 12$. Using the formula: contributions from each $k:$ $k=3$: $3*2^2*1= 3*4=12$. $k=1$ and $k=2$ contribute $0$. So total sum $12$, which matches.

\end{tcolorbox}

\end{tcolorbox}
\end{adjustbox}
\caption{Prompt baselines} 
\label{fig:prompt_baselines}
\end{figure}

\clearpage

\newpage

\section{Training of step-taggers} \label{sec:appendix-training-details}

\begin{figure}[h]
    \vspace{-0.2cm}
    \centering
    \begin{minipage}{0.49\linewidth}
        \includegraphics[width=\linewidth]{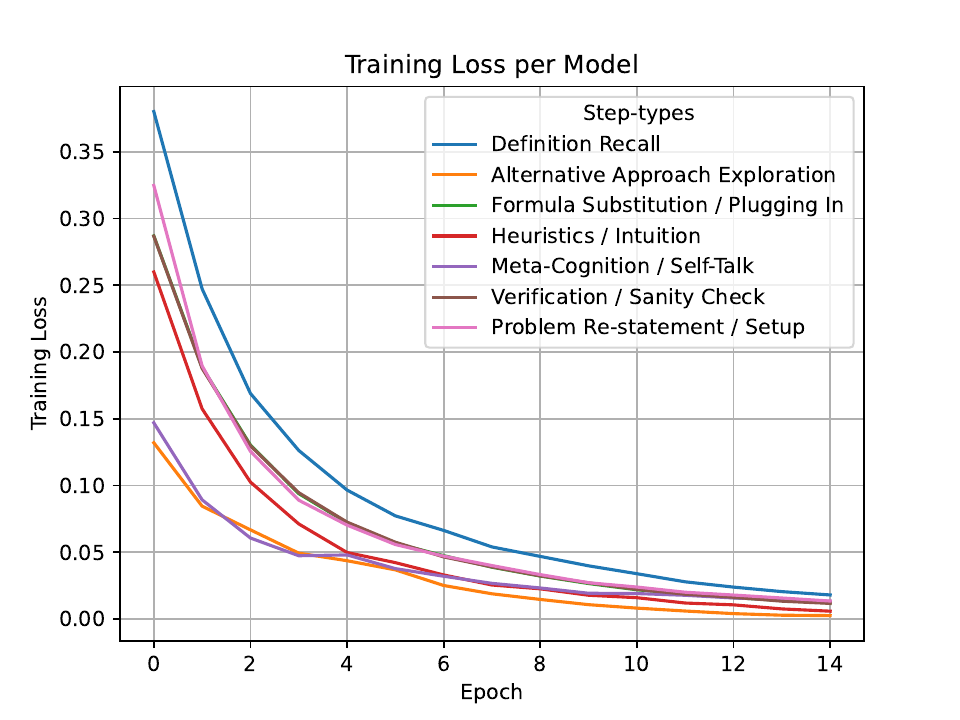}
        \subcaption{Training loss}
        \label{fig:loss}
    \end{minipage}
    \hspace{-0.05cm}
    \begin{minipage}{0.49\linewidth} 
        \includegraphics[width=\linewidth]{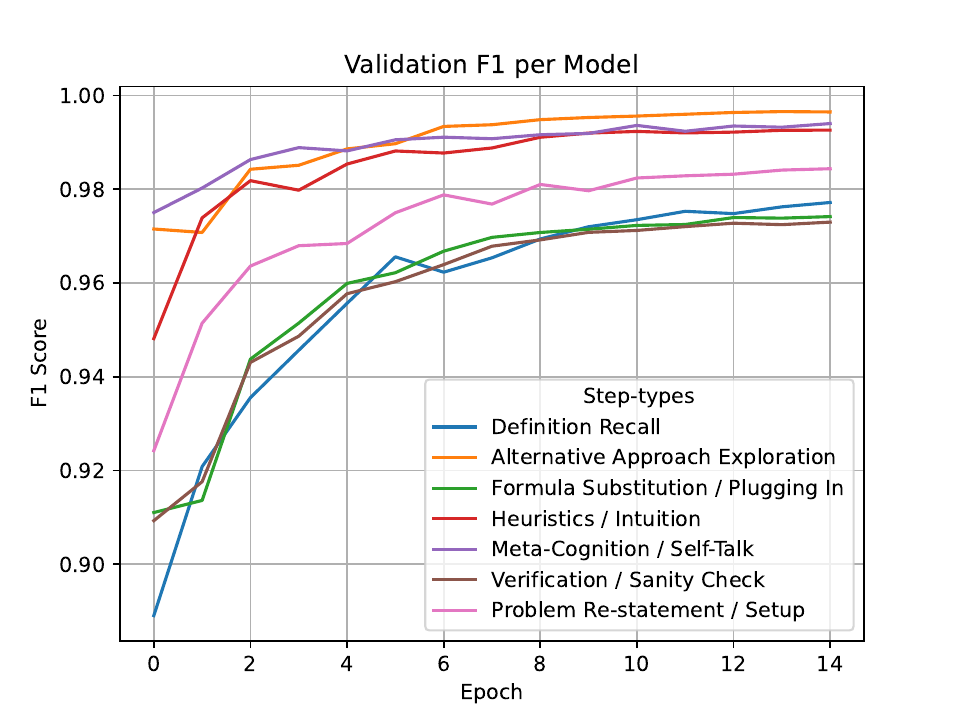}
        \subcaption{Validation Micro-F1}
        \label{fig:validation}
    \end{minipage}
    \vspace{-0.2cm}
    \caption{Training metrics of the BERT Step-Taggers}
    \label{fig:training-metrics}
    \vspace{-0.2cm}
\end{figure}

Figure \ref{fig:training-metrics} details the training metrics of our BERT step-taggers. Each step-tagger is trained to identify a specific step-type. We trained and validated our BERT models on reasoning traces obtained using DS-Qwen14B inferred on MATH500 and GPQA train datasets. We used a \emph{balanced cross-entropy} to enhance the performance of the models on low-represented classes. We trained classifiers for 15 epochs, with a batch size of 16 and we used an AdamW optimizer with a learning rate of $2.10^{-5}$.

\newpage

\section{Training-Inference cost trade-off of ST-ES} \label{sec:cost-trade-off}

This section presents an analysis of the cost of ST-ES. In this paper, we trained a seven BERT models (on seven different step-types). However, the data generation and labeling process is a one-time exercise, using the DS-Qwen14B on MATH500 and GPQA train split. Then, each step-tagger is used for all models and datasets configurations.

\begin{figure}[h]
    \centering
    \includegraphics[width=1\linewidth]{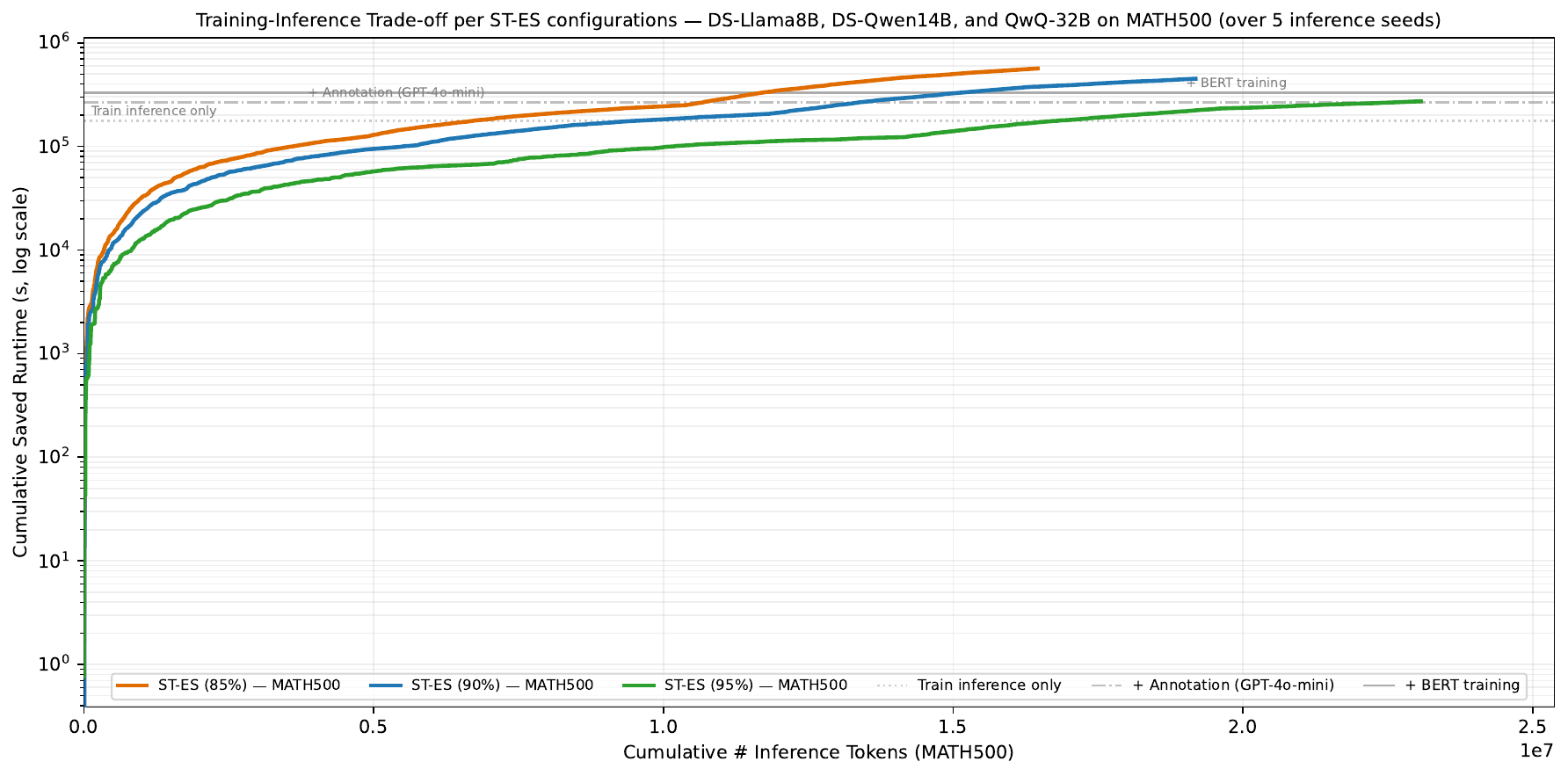}
    \caption{Cumulative saved runtime of ST-ES vs. Cumulative number of tokens generated on MATH500 test dataset for DS-Llama8B, DS-Qwen14B, and QwQ-32B. The gray lines represents the training costs of the seven BERT Step-tagging modules. We observe that configurations ST-ES 85\% and 90\% fully recovers the training costs of the BERT Step-Tagging modules.}
    \label{fig:cost-trade-off}
\end{figure}

Figure \ref{fig:cost-trade-off} presents the training-inference cost trade-off of ST-ES on the MATH500 test dataset, over the 3 selected models, across the five seeds for each models (to observe how ST-ES scales). We compare the estimated saved runtime of ST-ES against the training runtime. 

ST-ES testing inference time includes three components: (1) the runtime to generate the early-stopped trace (assuming linear relation between the token-count and the runtime \cite{oh2022improvingtopkdecodingnonautoregressive}), (2) the runtime to infer the BERT Step-Tagging module, and (3) the runtime to forcing the module to answer. 

The training runtime also includes three components: (1) the inference of DS-Qwen14B on training samples (MATH500 and GPQA train), (2) the annotation of reasoning steps obtained from the training inference, and (3) the training runtime of the BERT Step-Taggers (seven in total, one for each selected step-type).

We observe that $85\%$ and $90\%$ fully recovers the training costs solely with their inferences on the MATH500 dataset, and $95\%$ almost recovers all training costs. Furthermore, 85\%  saves almost two time the training runtime, while needing lower amount of tokens. But this is at the cost of the accuracy. 

We note that the gains expected by ST-ES would be even larger when including saved runtime on other datasets used in this paper (namely GSM8K, MMLU, and AIME). Importantly, the ST-ES early-stopping criteria is designed to be used at scale, and only requires a one-time training exercice.

\newpage

\section{Impact of BERT Step-Tagging performance on ST-ES} \label{sec:impact-bert-step-tagging}

\begin{figure*}[ht]
    \centering
    \begin{subfigure}{0.33\linewidth}
        \centering
        \includegraphics[width=\linewidth]{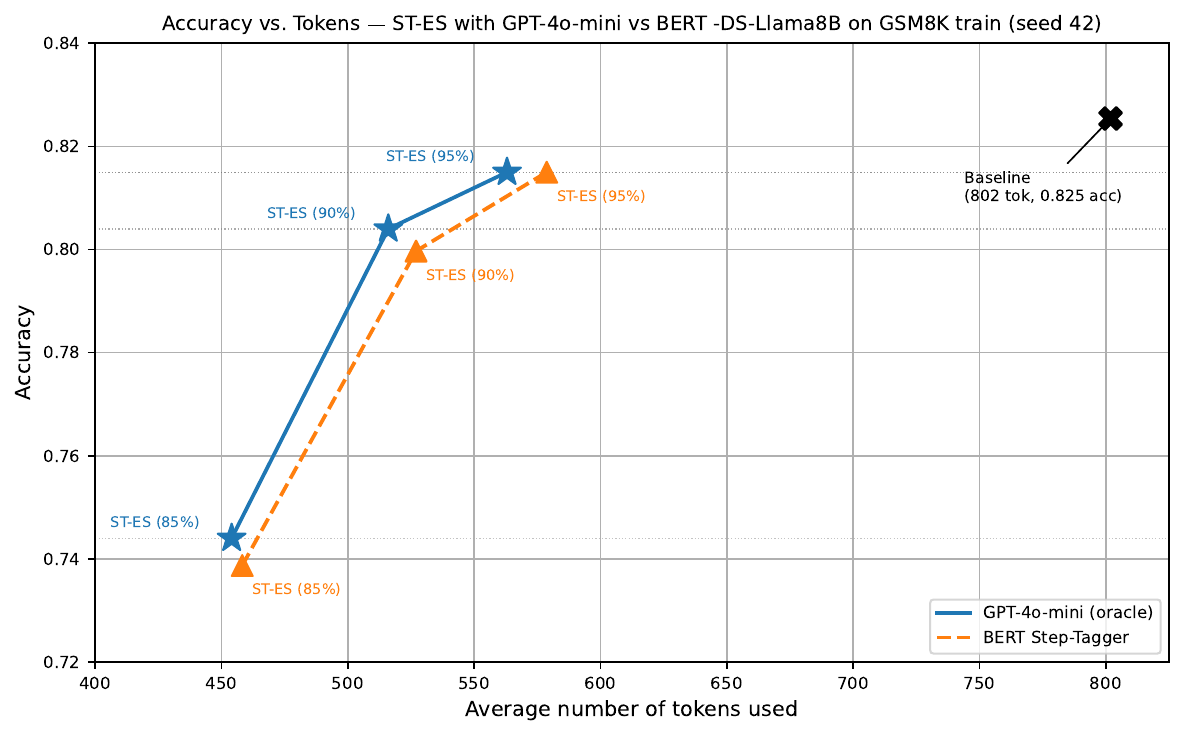}
        \caption{DS-Llama8B on GSM8K}
        \label{fig:DS8B-GSM8K-BERT-approx} 
    \end{subfigure}
    \hfill
    \begin{subfigure}{0.33\linewidth}
        \centering
        \includegraphics[width=\linewidth]{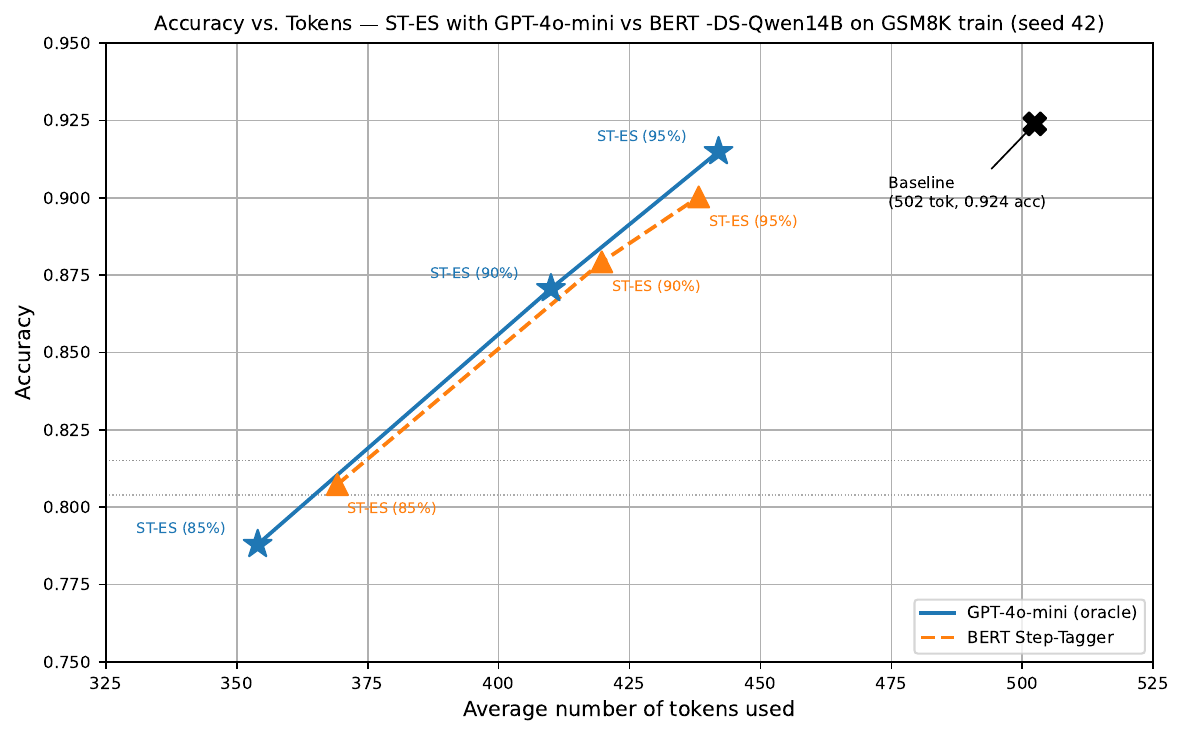}
        \caption{DS-Qwen14B on GSM8K}
        \label{fig:DS14B-GSM8K-BERT-approx}
    \end{subfigure}
    \hfill
    \begin{subfigure}{0.33\linewidth}
        \centering
        \includegraphics[width=0.99\linewidth]{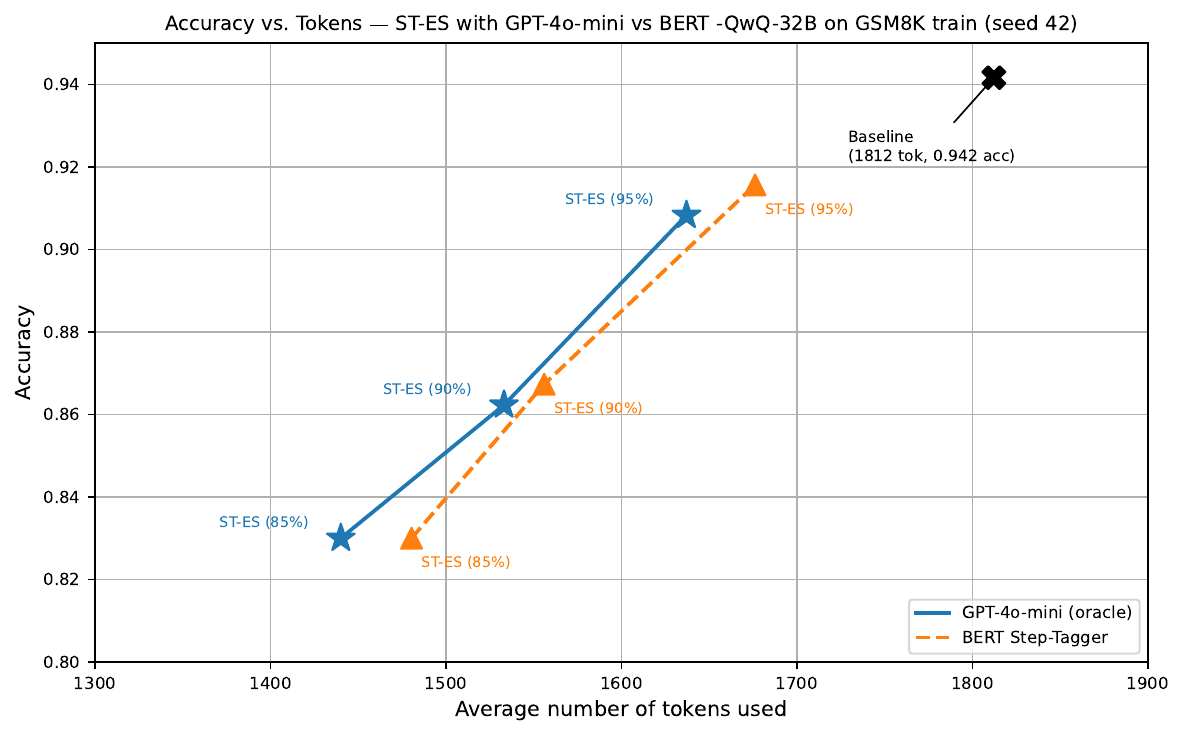}
        \caption{QwQ-32B on GSM8K}
        \label{fig:QwQ32B-GSM8K-BERT-approx}
    \end{subfigure}
    \begin{subfigure}{0.33\linewidth}
        \centering
        \includegraphics[width=\linewidth]{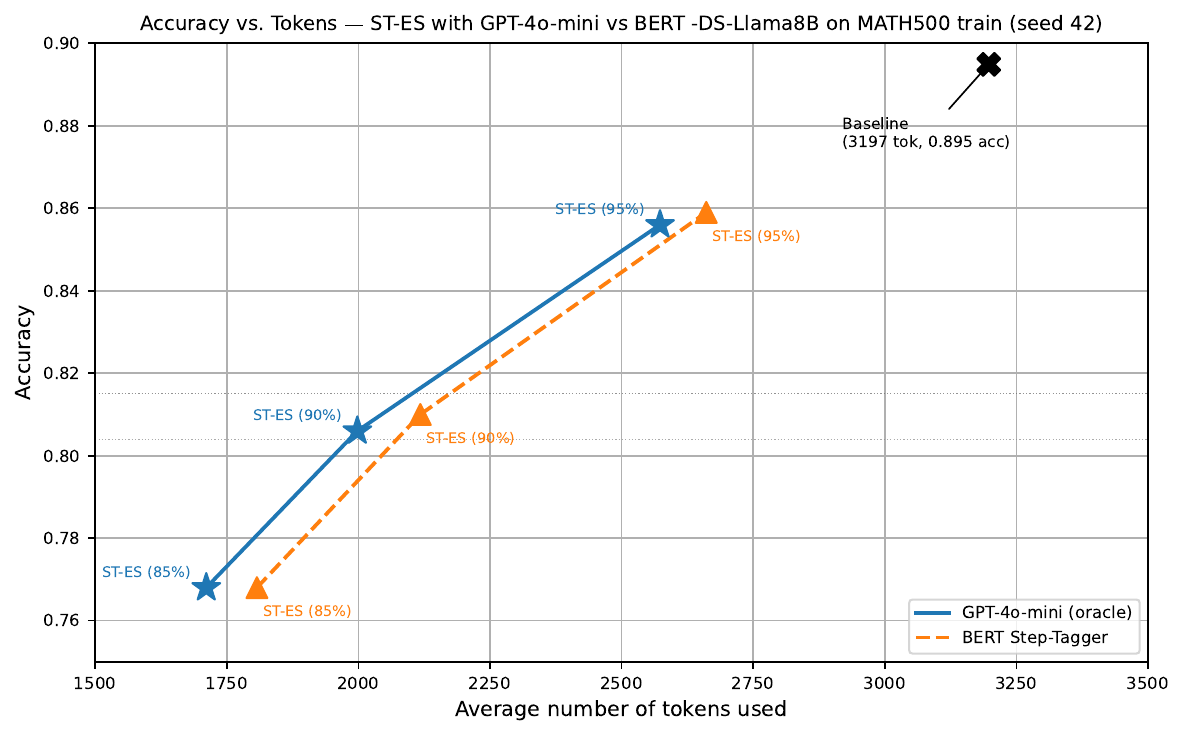}
        \caption{DS-Llama8B on MATH500}
        \label{fig:DS8B-MATH500-BERT-approx} 
    \end{subfigure}
    \hfill
    \begin{subfigure}{0.33\linewidth}
        \centering
        \includegraphics[width=\linewidth]{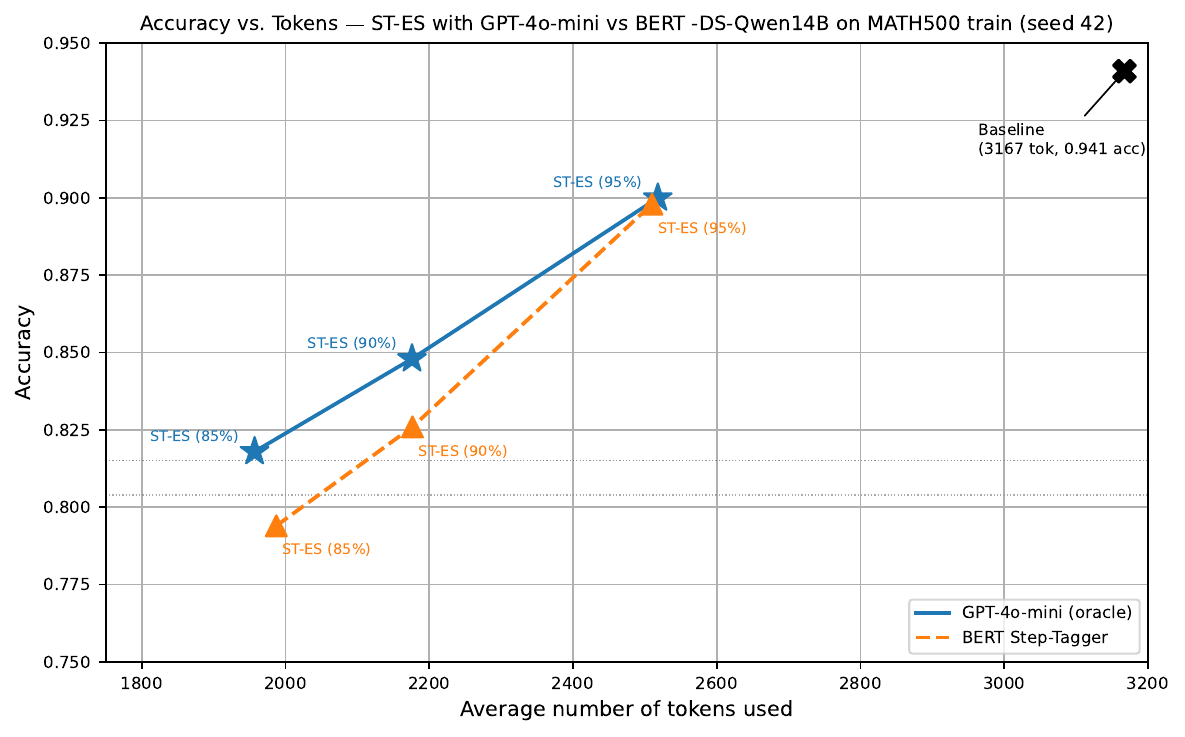}
        \caption{DS-Qwen14B on MATH500}
        \label{fig:DS14B-MATH500-BERT-approx}
    \end{subfigure}
    \hfill
    \begin{subfigure}{0.33\linewidth}
        \centering
        \includegraphics[width=0.99\linewidth]{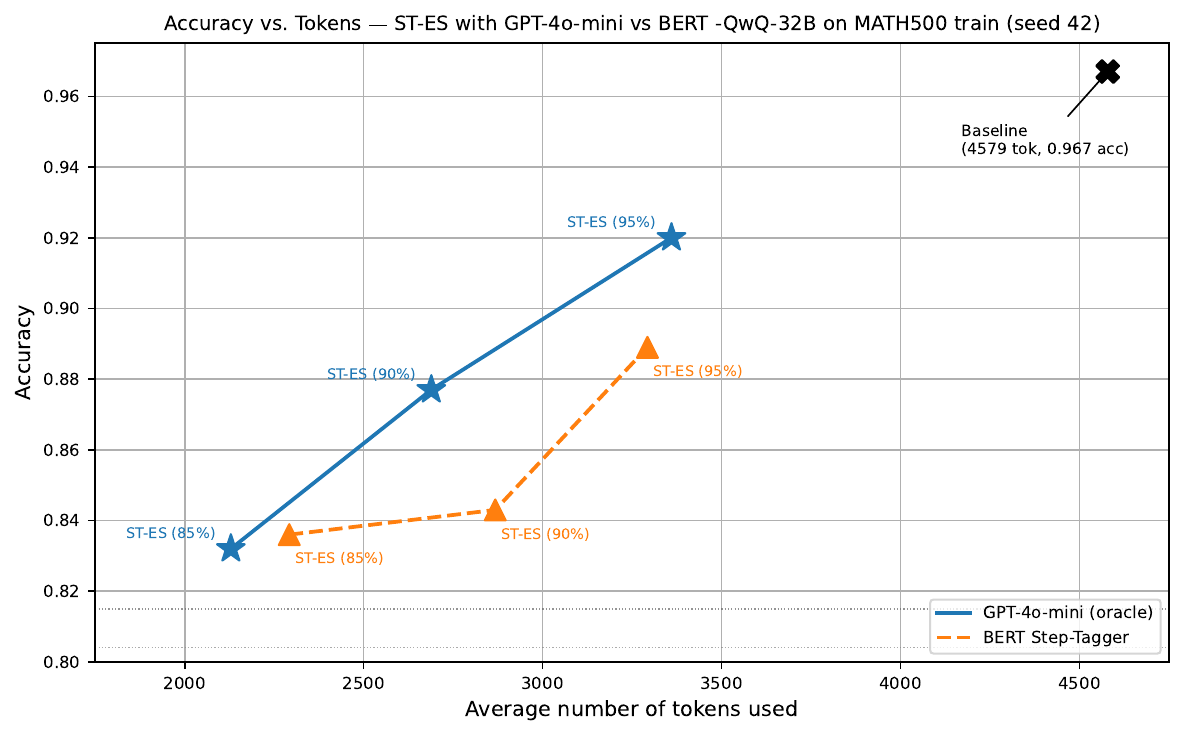}
        \caption{QwQ-32B on MATH500}
        \label{fig:QwQ32B-MATH500-BERT-approx}
    \end{subfigure}
    \\
    \begin{subfigure}{0.33\linewidth}
        \centering
        \includegraphics[width=\linewidth]{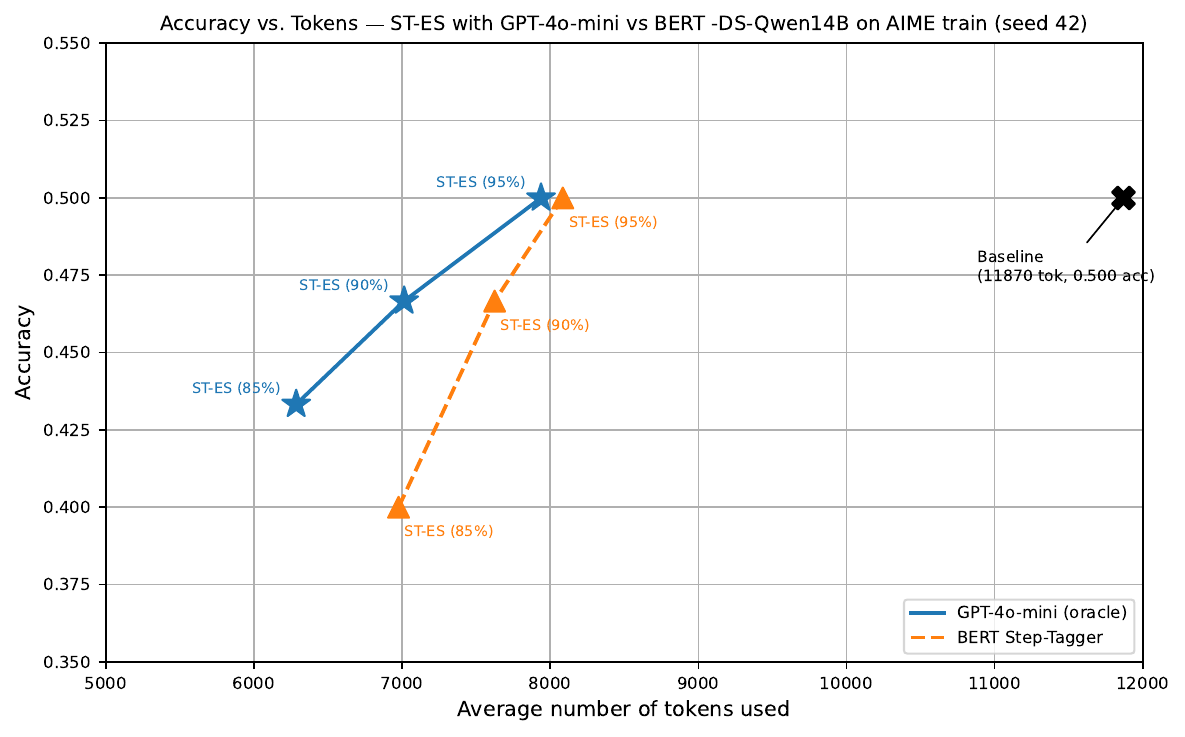}
        \caption{\small DS-Qwen14B on AIME}
        \label{fig:DS14B-AIME-BERT-approx} 
    \end{subfigure}
    \hfill
    \begin{subfigure}{0.33\linewidth}
        \centering
        \includegraphics[width=\linewidth]{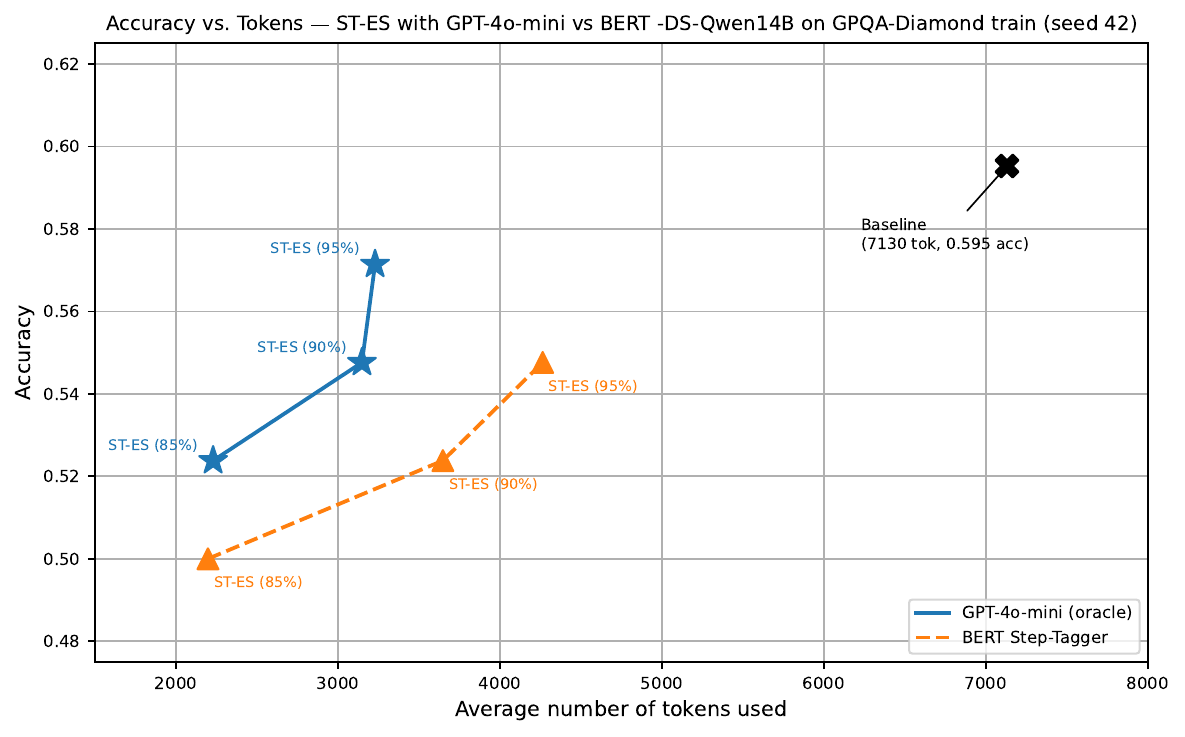}
        \caption{\small DS-Qwen14B on GPQA}
        \label{fig:DS14B-GPQA-BERT-approx}
    \end{subfigure}
    \hfill
    \begin{subfigure}{0.33\linewidth}
        \centering
        \includegraphics[width=\linewidth]{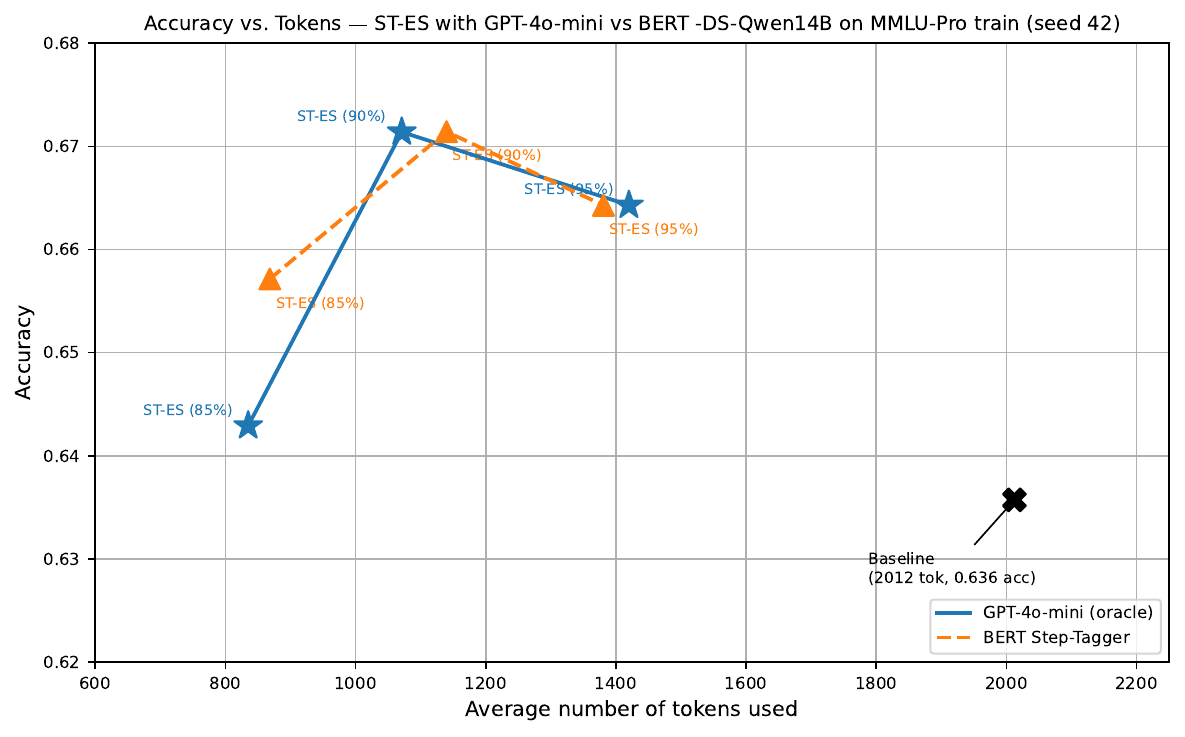}
        \caption{\small DS-Qwen14B on MMLU-Pro}
        \label{fig:DS14B-MMLU-BERT-approx}
    \end{subfigure}
    \vspace{-0.2cm}
    \caption{\small Number of Tokens vs. Pass@$1$ - ST-ES with \texttt{GPT-4o-mini} vs. BERT Step-tagging on train datasets - seed 42}
    \label{fig:st-es-BERT-approx}
    \vspace{-0.5cm}
\end{figure*}

Figure \ref{fig:st-es-BERT-approx} compares ST-ES criteria applied using the \texttt{GPT-4o-mini} (in blue) and BERT Step-Tagging (in orange) tags. We observe that the approximation achieved best results on GSM8K and MATH500 datasets. On AIME, GPQA, and MMLU, the difference in performance is higher, but still acceptable (less than 2\% accuracy drop, and 5 to 10\% higher token-count). We suspect that this is due to two reasons. First, the higher value of $\delta$ on harder tasks means that if the BERT taggers are making an error, this error is likely to be propagated across longer reasoning chains. Second, the datasets contains much less samples, future work should look at expanding the datasets to observe how the results scales.

\newpage

\section{Evaluation of Step-Tagging Early-Stopping} \label{sec:appendix-st-es-performance}


Table \ref{tab:full_results_5_seeds} reports all the token-usage, the proportion of saved number of tokens, the Avg@5, the Pass@5 and the Cons@5 for all configurations. Results are averaged over the 5 seeds we used. 

\begin{table}[h]
\centering
\tiny
\begin{tabular}{ll
                ccccc
                ccccc}
\toprule
\multirow{2}{*}{\textbf{Model}} & \multirow{2}{*}{\textbf{Config.}} 
& \multicolumn{5}{c}{\textbf{MATH500}} 
& \multicolumn{5}{c}{\textbf{GSM8K}} \\
\cmidrule(lr){3-7} \cmidrule(lr){8-12}
& & \# Tokens & Saved (\%) & Avg@5 & Pass@5 & Cons@5 & \# Tokens & Saved (\%) & Avg@5 & Pass@5 & Cons@5 \\
\midrule

\multirow{10}{*}{DS-8B}
  & Standard & 3655.0 & -- & 0.878 & 0.970 & 0.726 & 958.3 & -- & 0.829 & 0.943 & 0.651 \\
  
  \cmidrule(lr){3-7} \cmidrule(lr){8-12}
  
  & Basel. $\mathcal{P}^{(0)}_{\text{user}}$ & 2989.6 & 18.21 & 0.866 & 0.952 & 0.722 & 525.8 & 45.13 & 0.771 & 0.917 & 0.579 \\
  & Basel. $\mathcal{P}^{(0)}_{\text{system}}$ & 2634.4 & 27.92 & 0.817 & 0.960 & 0.592 & 456.9 & 52.32 & 0.763 & 0.895 & 0.574 \\
  & Basel. $\mathcal{P}^{(1)}_{\text{system}}$ & 2139.5 & 41.46 & 0.782 & 0.942 & 0.526 & 560.8 & 41.48 & 0.754 & 0.914 & 0.537 \\
  & Basel. $\mathcal{P}^{(3)}_{\text{system}}$ & 2565.3 & 29.81 & 0.789 & 0.952 & 0.540 & 830.5 & 13.34 & 0.748 & 0.904 & 0.541 \\
  
  \cmidrule(lr){3-7} \cmidrule(lr){8-12}
  
  & ST-ES ($85\%$) & 1801.4 & 50.71 & 0.761 & 0.908 & 0.568 & 521.5 & 45.58 & 0.708 & 0.842 & 0.575 \\
  & ST-ES ($90\%$) & 2120.3 & 41.99 & 0.793 & 0.920 & 0.616 & 618.8 & 35.43 & 0.788 & 0.884 & 0.692 \\
  & ST-ES ($95\%$) & 2614.5 & 28.47 & 0.845 & 0.936 & 0.704 & 696.7 & 27.29 & 0.809 & 0.890 & 0.727 \\
  
\midrule

\multirow{10}{*}{DS-14B}
  & Standard & 3388.8 & -- & 0.923 & 0.980 & 0.836 & 662.9 & -- & 0.910 & 0.952 & 0.843 \\
  
  \cmidrule(lr){3-7} \cmidrule(lr){8-12}
  
  & Basel. $\mathcal{P}^{(0)}_{\text{user}}$ & 2691.5 & 20.58 & 0.933 & 0.982 & 0.834 & 505.1 & 23.80 & 0.856 & 0.956 & 0.662 \\
  & Basel. $\mathcal{P}^{(0)}_{\text{system}}$ & 2346.2 & 30.77 & 0.886 & 0.966 & 0.754 & 470.9 & 28.96 & 0.873 & 0.949 & 0.710 \\
  & Basel. $\mathcal{P}^{(1)}_{\text{system}}$ & 2211.4 & 34.74 & 0.873 & 0.974 & 0.708 & 566.5 & 14.54 & 0.838 & 0.952 & 0.629 \\
  & Basel. $\mathcal{P}^{(3)}_{\text{system}}$ & 2535.0 & 25.19 & 0.879 & 0.968 & 0.748 & 839.6 & -26.65 & 0.841 & 0.952 & 0.631 \\
  
  \cmidrule(lr){3-7} \cmidrule(lr){8-12}
  
  & ST-ES ($85\%$) & 2158.4 & 36.30 & 0.827 & 0.950 & 0.646 & 377.2 & 43.09 & 0.527 & 0.802 & 0.262 \\
  & ST-ES ($90\%$) & 2393.6 & 29.36 & 0.853 & 0.960 & 0.688 & 415.9 & 37.26 & 0.621 & 0.843 & 0.368 \\
  & ST-ES ($95\%$) & 2851.9 & 15.84 & 0.909 & 0.972 & 0.812 & 464.6 & 29.91 & 0.865 & 0.946 & 0.723 \\

\midrule

\multirow{10}{*}{QwQ-32B}
  & Standard & 4475.3 & -- & 0.954 & 0.984 & 0.898 & 2075.7 & -- & 0.953 & 0.965 & 0.934 \\
  
  \cmidrule(lr){3-7} \cmidrule(lr){8-12}
  
  & Basel. $\mathcal{P}^{(0)}_{\text{user}}$ & 2908.8 & 35.00 & 0.955 & 0.986 & 0.916 & 988.0 & 52.40 & 0.952 & 0.968 & 0.937  \\
  & Basel. $\mathcal{P}^{(0)}_{\text{system}}$ & 3201.1 & 28.47 & 0.932 & 0.976 & 0.852 & 833.3 & 59.85 & 0.940 & 0.974 & 0.869 \\
  & Basel. $\mathcal{P}^{(1)}_{\text{system}}$ & 3182.4 & 28.89 & 0.925 & 0.974 & 0.856 & 871.2 & 58.02 & 0.943 & 0.975 & 0.876 \\
  & Basel. $\mathcal{P}^{(3)}_{\text{system}}$ & 3665.5 & 18.09 & 0.926 & 0.974 & 0.858 & 1387.3 & 33.16 & 0.935 & 0.974 & 0.855 \\
  
  \cmidrule(lr){3-7} \cmidrule(lr){8-12}
  
  & ST-ES ($85\%$) & 2415.1 & 46.03 & 0.839 & 0.950 & 0.666 & 1152.1 & 44.49 & 0.833 & 0.945 & 0.646  \\
  & ST-ES ($90\%$) & 2953.9 & 33.99 & 0.889 & 0.960 & 0.758 & 1338.1 & 35.53 & 0.899 & 0.958 & 0.794  \\
  & ST-ES ($95\%$) & 3559.5 & 20.46 & 0.922 & 0.966 & 0.846 & 1462.2 & 29.55 & 0.943 & 0.964 & 0.903  \\

\bottomrule
\end{tabular}
\caption{Performance of Step-Tagging Early stopping - 5 seeds (40, 41, 42, 43, 44)}
\label{tab:full_results_5_seeds}
\end{table}


Table \ref{tab:other_dataset_st_es} shows the token-usage, the proportion of saved number of tokens, and the Pass@1 for DS-Qwen14B on AIME, GPQA, and MMLU test datasets. Results are obtained with a seed of $42$.

\begin{table}[h]
\centering
\tiny
\begin{tabular}{l
                ccc
                ccc
                ccc}
\toprule
\multirow{2}{*}{\textbf{Config.}} 
& \multicolumn{3}{c}{\textbf{AIME 23-24}} & \multicolumn{3}{c}{\textbf{GPQA-Diamond}} & \multicolumn{3}{c}{\textbf{MMLU-Pro}} \\
\cmidrule(lr){2-4} \cmidrule(lr){5-7} \cmidrule(lr){8-10}
& \# Tokens & Saved (\%) & Pass@1 & \# Tokens & Saved (\%) & Pass@1 & \# Tokens & Saved (\%) & Pass@1 \\
\midrule

Standard & 13096.3 & - & 0.483 & 7024.7 & - & 0.603 & 2501.9 & - & 0.652 \\
  
  \cmidrule(lr){2-10} 
  
Basel. $\mathcal{P}^{(0)}_{\text{user}}$ & 12748.4 & 2.66 & 0.422 & 4801.0 & 31.66 & 0.577 & 2194.1 & 12.30 & 0.665 \\
Basel. $\mathcal{P}^{(0)}_{\text{system}}$ & 11095.9 & 15.27 & 0.500 & 4768.3 & 32.12 & 0.571 & 2292.9 & 8.35 & 0.675 \\
Basel. $\mathcal{P}^{(1)}_{\text{system}}$ & 10181.2 & 22.26 & 0.517 & 3736.2 & 46.81 & 0.539 & 1705.6 & 31.83 & 0.637 \\
Basel. $\mathcal{P}^{(3)}_{\text{system}}$ & 11635.7 & 11.15 & 0.467 & 4530.1 & 35.51 & 0.564 & 1935.2 & 22.65 & 0.613 \\
  
  \cmidrule(lr){2-10} 
  
ST-ES ($85\%$) & 6082.9 & 53.55 & 0.283 & 2383.2 & 66.07 & 0.487 & 927.6 & 62.92 & 0.606 \\
ST-ES ($90\%$) & 8609.6 & 34.26 & 0.383 & 3326.6 & 52.64 & 0.557 & 1153.1 & 53.91 & 0.643 \\
ST-ES ($95\%$) & 9299.8 & 28.99 & 0.417 & 4432.9 & 36.89 & 0.564 & 1469.6 & 41.26 & 0.670 \\

\bottomrule
\end{tabular}
\caption{Performance of ST-ES - DS-Qwen14B - seed 42}
\label{tab:other_dataset_st_es}
\end{table}


\end{document}